\documentclass[letterpaper, 10 pt, conference]{ieeeconf}  

\IEEEoverridecommandlockouts                              

\usepackage{booktabs}
\usepackage{cuted}
\usepackage{soul}
\usepackage{hyperref}
\usepackage{verbatimbox}
\usepackage{graphicx}
\usepackage[export]{adjustbox}
\usepackage{amsmath}
\usepackage{todonotes}
\usepackage{xcolor}
\usepackage{afterpage}%
\usepackage{array}
\usepackage{caption,setspace}
\usepackage{bm}
\usepackage{subcaption}
\newcommand{\secref}[1]{Sec.~\ref{#1}}
\newcommand{\figref}[1]{Fig.~\ref{#1}}
\usepackage{letltxmacro}
\usepackage{amsfonts} 
\LetLtxMacro{\originaleqref}{\eqref}
\newcommand{\tabref}[1]{Tab.~\ref{#1}}
\renewcommand{\eqref}{Eq.~\originaleqref}
\newcommand\myworries[1]{\textcolor{black}{#1}}
\definecolor{tablecolor}{RGB}{0,0,0}
\title{\LARGE \bf
Uncertainty-aware Panoptic Segmentation
}

\author{Kshitij Sirohi$^{1}$, Sajad Marvi$^{1}$, Daniel B\"uscher$^{1}$ and Wolfram Burgard$^{2}$
\thanks{
$^{1}$Department of Computer Science, University of Freiburg, Germany.
$^{2}$Department of Engineering, Technical University Nürnberg, Germany.
This work was financed by the Baden-Württemberg Stiftung gGmbH.
}%
}
\begin{document}
\maketitle
\thispagestyle{empty}
\pagestyle{empty}

\begin{abstract}

Reliable scene understanding is indispensable for modern autonomous systems. Current learning-based methods typically try to maximize their performance based on segmentation metrics that only consider the quality of the segmentation. However, for the safe operation of a system in the real world it is crucial to consider the uncertainty in the prediction as well. In this work, we introduce the novel task of uncertainty-aware panoptic segmentation, which aims to predict per-pixel semantic and instance segmentations, together with per-pixel uncertainty estimates. We define two novel metrics to facilitate its quantitative analysis, the uncertainty-aware Panoptic Quality (uPQ) and the panoptic Expected Calibration Error (pECE). We further propose the novel top-down Evidential Panoptic Segmentation Network (EvPSNet) to solve this task. Our architecture employs a simple yet effective panoptic fusion module that leverages the predicted uncertainties. 
Furthermore, we provide several strong baselines combining state-of-the-art panoptic segmentation networks with sampling-free uncertainty estimation techniques. Extensive evaluations show that our EvPSNet achieves the new state-of-the-art for the standard Panoptic Quality (PQ), as well as for our uncertainty-aware panoptic metrics.
We make the code available at: \url{https://github.com/kshitij3112/EvPSNet}
\end{abstract}


\section{Introduction}

Due to the recent advances in deep learning,
perception systems of modern autonomous systems largely rely on
convolutional neural networks (CNNs), in particular for the tasks
of semantic segmentation \cite{lateef2019survey} and object detection \cite{liu2020deep}, utilizing different sensors \cite{vertens2020heatnet,zurn2022self}.
However, these two similar tasks are still often treated separately.
Aiming for a holistic scene understanding, Kirillov \emph{et al.} \cite{kirillov2019panoptic} introduced
panoptic segmentation for combined segmentation of \textit{stuff} classes,
consisting of amorphous regions like road surfaces,
and \textit{thing} classes, consisting of distinct instances of objects like cars and pedestrians. 

\begin{figure}
\captionsetup[subfigure]{aboveskip=-2.5ex,belowskip=1ex,font={footnotesize,color=white}}
\begin{subfigure}{0.495\linewidth}
 \includegraphics[width=1.0\linewidth]{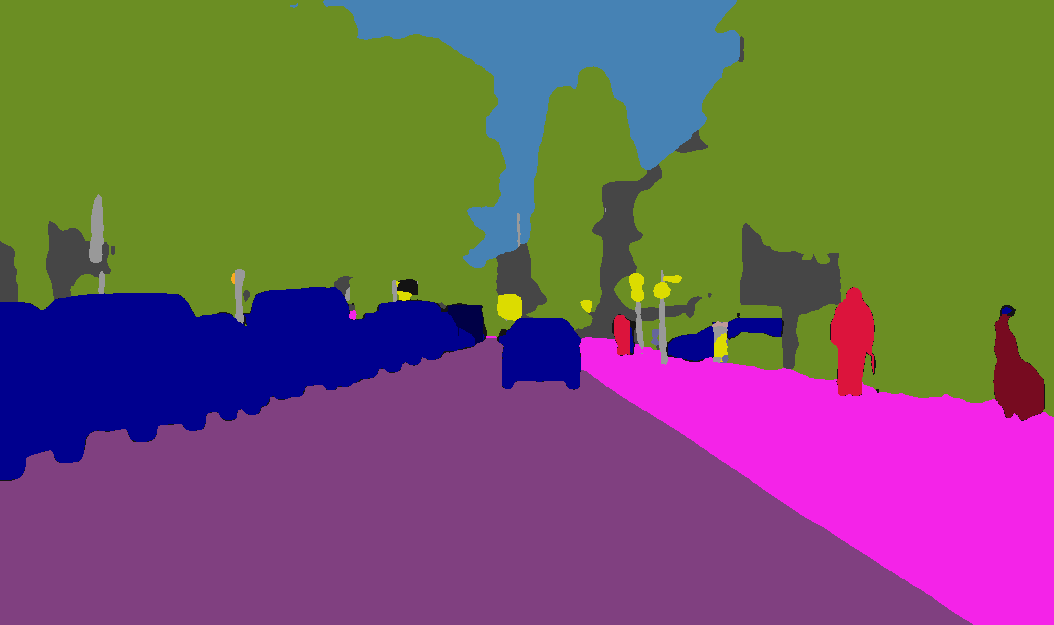}
 \subcaption{semantic segmentation}\label{fig:sem_seg}
\end{subfigure}
\begin{subfigure}{0.495\linewidth}
 \includegraphics[width=1.0\linewidth]{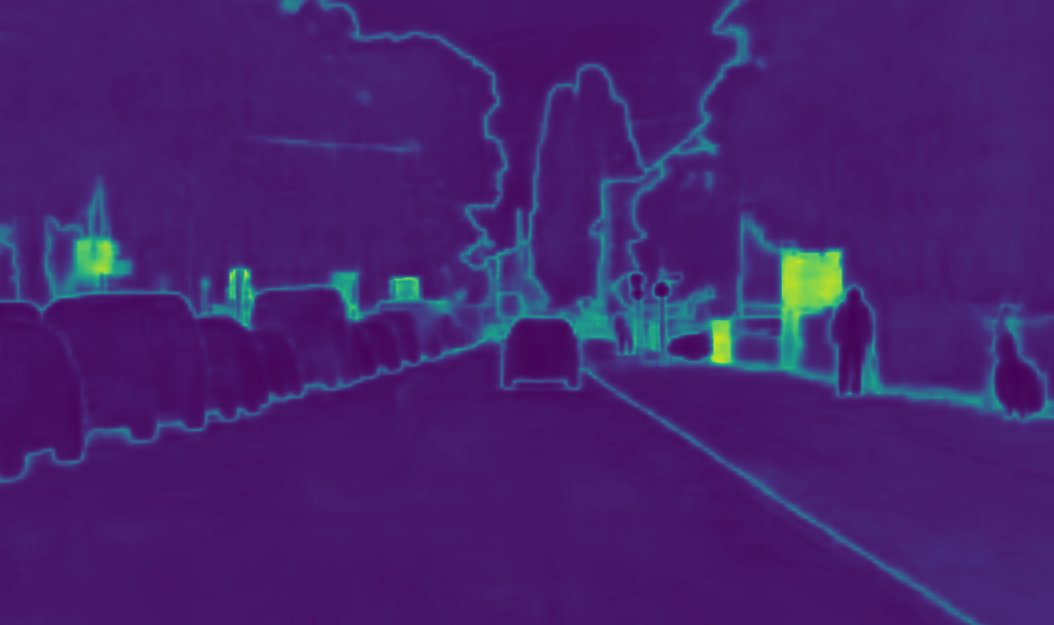}
 \subcaption{semantic uncertainties}\label{fig:sem_unc}
\end{subfigure}
\begin{subfigure}{0.495\linewidth}
 \includegraphics[width=1.0\linewidth]{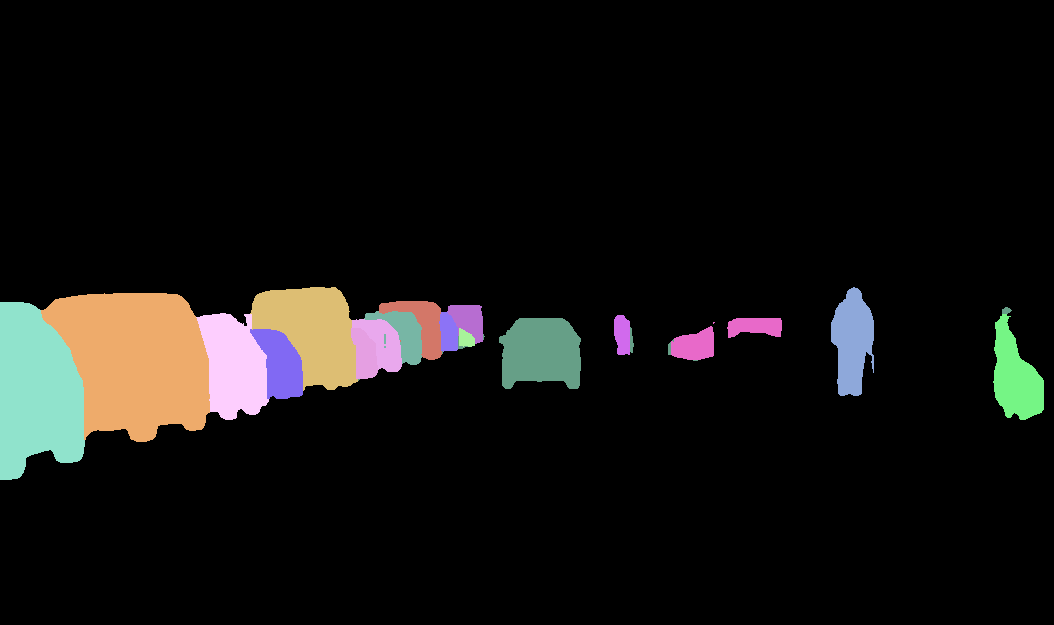}
 \subcaption{instance segmentation}\label{fig:instance}
\end{subfigure}
\begin{subfigure}{0.495\linewidth}
 \includegraphics[width=1.0\linewidth]{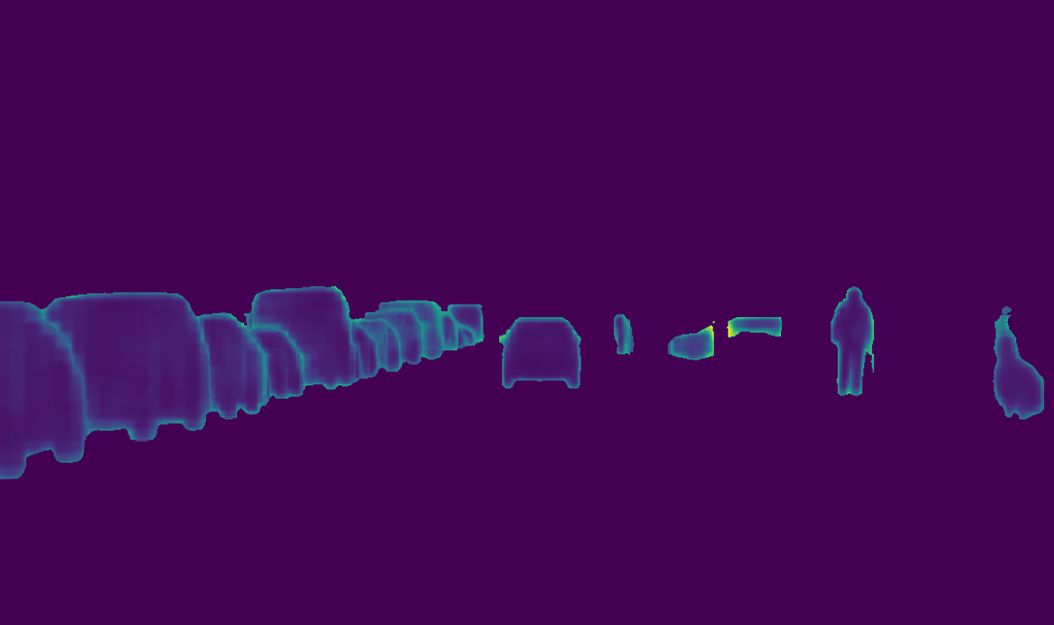}
 \subcaption{instance uncertaintites}\label{fig:inst_unc}
\end{subfigure}
\begin{subfigure}{0.495\linewidth}
 \includegraphics[width=1.0\linewidth]{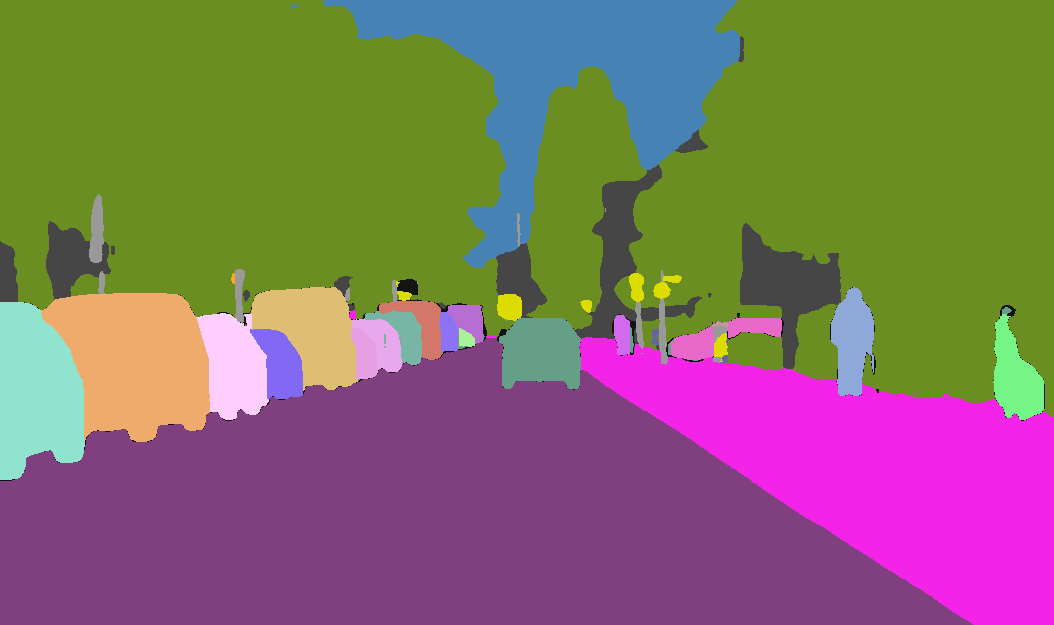}
 \subcaption{panoptic segmentation}\label{fig:panoptic}
\end{subfigure}
\begin{subfigure}{0.495\linewidth}
 \includegraphics[width=1.0\linewidth]{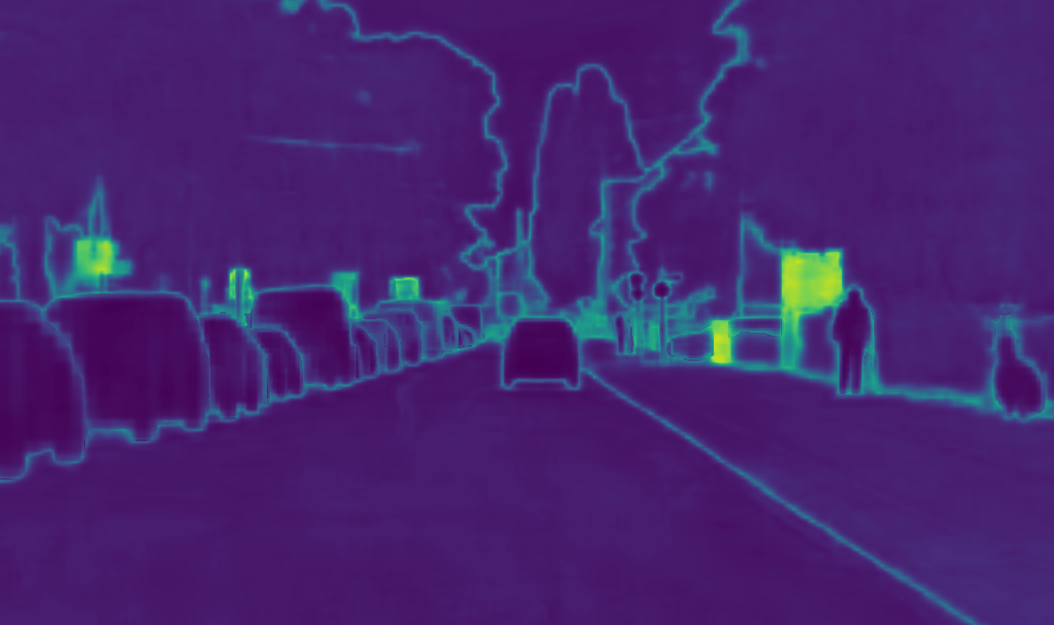}
 \subcaption{panoptic uncertaintites}\label{fig:pan_unc}
\end{subfigure}
\caption{Semantic, instance and panoptic segmentations with their associated uncertainties as predicted by our EvPSNet for the Cityscapes data.}
\label{fig:examples}
\end{figure}


Several existing panoptic segmentation methods provide CNN-based architectures for different modalities, such as cameras and LiDARs \cite{mohan2021efficientps,sirohi2021efficientlps}.
Being supervised learning-based approaches, these networks first learn on a training dataset,
and then evaluate their performance using specific metrics on a test set.
A typical limitation is that the training and test sets usually have quite similar data distributions and conditions,
while these can be quite different during deployment in the real world consisting of objects not seen during training.
Since current training metrics typically consider only the performance,
the resulting networks can be quite overconfident in their (false) predictions,
possibly posing safety-critical threats in autonomous driving scenarios.
In general, these approaches lack insight
into the performance of the network in unseen environments and
into the reliability of it's confidence estimate output.

In this work, we introduce the task of uncertainty-aware panoptic segmentation:
Panoptic segmentation accompanied by calibrated per-pixel uncertainty estimates, covering both the semantic and instance uncertainties.
This task,
illustrated in \figref{fig:examples}, intends to
provide reliable predictions even in challenging scenarios and to
motivate future research in the field of holistic scene understanding.
Conventional methods utilize the simple softmax operation to provide probability estimates, which are quite limited in their reliability and typically inflated \cite{sensoy2018evidential}.
On the other hand, sampling-based methods such as dropout are primary candidates employed for reliable uncertainty estimation in various tasks.
However, these approaches are computationally intense and thus are not suitable for real-time applications, such as autonomous driving.
In this context the research in sampling-free methods for uncertainty estimation is gaining interest.
One such method is evidential deep learning,
which is already being used successfully in classification \cite{sensoy2018evidential},
regression \cite{amini2020deep}, and multitask learning settings \cite{petek2021robust}.

We propose a novel evidential panoptic segmentation network,
the EvPSNet architecture,
consisting of a shared backbone, separate uncertainty-aware semantic and instance segmentation heads, and a novel panoptic fusion module.
In this multitask learning setting, we utilize deep evidential learning for
simultaneously predicting semantic and instance segmentation outputs
and the corresponding pixel-wise uncertainties.
As a critical component, our intuitive fusion module
leverages the predicted uncertainties of both tasks
and provides the panoptic segmentation and uncertainty outputs.
We further introduce two new metrics, the uncertainty-aware Panoptic Quality (uPQ) and the panoptic Expected Calibration Error (pECE),
allowing to properly evaluate the uncertainty-aware panoptic segmentation
and the calibration of the uncertainty estimate, respectively.
We provide several strong sampling-free baselines to compare our method against.
In summary, our contributions are:

\begin{itemize}
 \item The definition of a novel simultaneous panoptic segmentation and uncertainty estimation task: the uncertainty-aware panoptic segmentation.
 \item The EvPSNet architecture, aiming to solve this task.
 \item An improved panoptic fusion module as part of EvPSNet, utilizing and propagating uncertainties.
 \item The uncertainty-aware Panoptic Quality (uPQ) and panoptic Expected Calibration Error (pECE) metrics.
 \item Several strong sampling-free baselines for comparison.
\end{itemize}

To the best of our knowledge, our approach is the first dedicated approach
to simultaneously estimate panoptic segmentation and associated per-pixel uncertainties.

\section{Related Work}

\subsection{Panoptic Segmentation}

Since the inception of the panoptic segmentation task, methods have generally taken one of two approaches:
either proposal-free (bottom-up) or proposal-based (top-down).
In the proposal-free approach \cite{cheng2020panoptic, wang2020axial} first a semantic segmentation is performed, followed by clustering the pixels belonging to the thing classes.
The corresponding clustering methods include center and offset regression \cite{cheng2020panoptic}, calculation of pixels affinity \cite{gao2019ssap}, or a Hough voting scheme \cite{leibe2004combined}.
In contrast, proposal-based methods \cite{kirillov2019panoptic2, mohan2021efficientps} consist of two parallel heads,
one to perform semantic segmentation and the other to predict bounding boxes and instance masks for thing class objects.
The instance segmentation part usually consists of a (modified) version of Mask-RCNN \cite{he2017mask}.
The final panoptic output is obtained by fusing the outputs of heads.
However, the separate heads lead to an overlapping mask problem due to the possible disagreement of semantic segmentation and instance segmentation results.
One way to deal with this is to learn the panoptic segmentation based on the semantic and instance logits
in a parameter-free fashion as a post-processing fusion technique \cite{mohan2021efficientps,xiong2019upsnet},
which, however, does not provide an estimate on the prediction uncertainty. Moreover, the proposed extensions to the panoptic segmentation task also do not provide any uncertainty estimation \cite{mohan2022amodal, gosala2022bird, de2021part}.



\subsection{Uncertainty Estimation}

The image classification task has been a popular topic among the research works on uncertainty estimation with neural networks for quite some time.
Bayesian neural networks (BNNs) learn the distribution over network weights to provide a probabilistic model for a network's output.
Bundell proposed utilizing Gaussian distribution over the network weights, representing each weight by mean and variance.
The result is an increase in the number of model parameters by a factor of two, hence it is not ideal for implementations with deep networks.
Gal \emph{et al.} \cite{gal2016dropout} proposed the method Monte Carlo (MC) dropout using dropout for variational inference.
Here, neurons are dropped during training, which can be interpreted as forming a Bernoulli distribution over the network weights.
The same image is fed to the network $N$ times with dropout during test time, and an uncertainty is calculated from these $N$ outputs.

The disadvantage of sampling-based methods is that they are not fit for real-time applications,
as they either need multiple passes through the network or multiple ensemble networks utilizing more computation resources.
Huang \emph{et al.} \cite{huang2018efficient} proposed using MC dropout to compute semantic segmentation uncertainty by taking the last $N$ frames from a video stream rather than passing one image $N$ times to save runtime.
However, this method is not able to predict uncertainty from a single image.

Guo \emph{et al.} \cite{guo2017calibration} proposed a sampling-free method called temperature scaling (TS), to learn a scaling factor for the learned logits, calibrating the predicted probabilities.
Recently, Li \emph{et al.} \cite{li2022uncertainty} adapted Radial Basis Functions Network (RBFN) \cite{van2020uncertainty} to provide uncertainty aware proposal segmentation with the aim to detect and predict uncertainties for out-of-distribution objects.
They predict the uncertainty for a particular class by utilizing the distance of feature vectors to their respective class centroids in the feature space.

Sensoy \emph{et al.} \cite{sensoy2018evidential} utilized evidential theory to introduce deep evidential learning to quantify classification uncertainty in a sampling-free fashion.
Here, the network is trained to collect parameters for a high-order distribution, the Dirichlet distribution in their case, from which the uncertainty of the prediction is computed.
Petek \emph{et al.} \cite{petek2021robust} utilized this method in multitask learning setting to simultaneously predict semantic segmentation and bounding box regression uncertainties.
In our approach, we build upon evidential deep learning to learn semantic segmentation, instance segmentation, and bounding box classification uncertainties.
We utilize the predicted uncertainties to fuse the instance and semantic segmentation outputs to achieve simultaneous panoptic segmentation and uncertainty estimation.

\section{Technical Approach}



\begin{figure}
\centering
\includegraphics[width=0.45\textwidth]{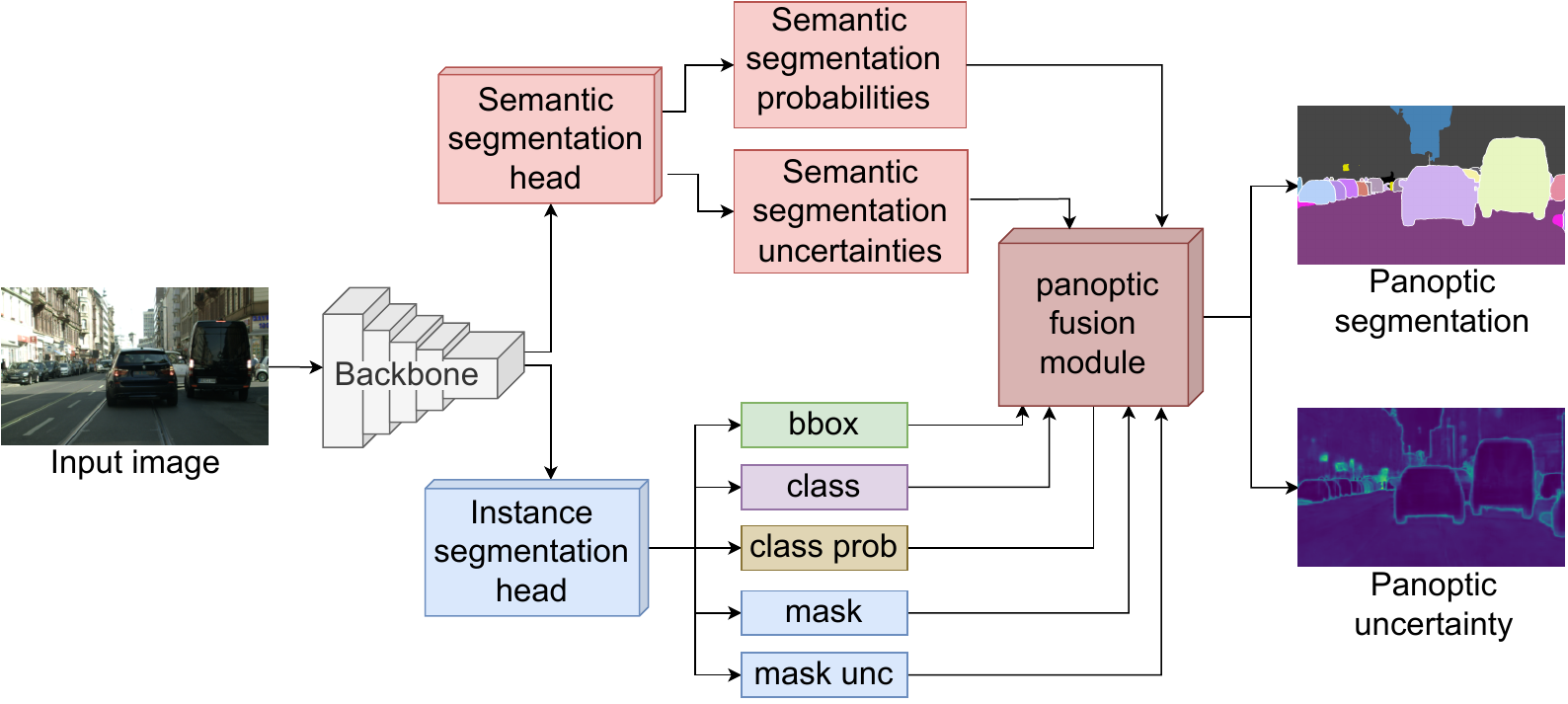}
\caption{
Overview of our proposed EvPSNet architecture. Our network utilizes deep evidential learning to provide semantic segmentation probabilities and uncertainties from the semantic head, as well as bounding boxes (bbox), class (cls), class probability (cls prob), mask, and mask uncertainties (mask unc) from the instance head. These are combined by our panoptic fusion module to provide panoptic segmentation and uncertainties. 
} 
\label{fig:architecture}
\end{figure}

\subsection{Network Architecture}

On overview of our network architecture is given in \figref{fig:architecture}.
We utilize a state-of-the-art proposal-based panoptic segmentation network, EfficientPS \cite{mohan2021efficientps}, as our base network.
It consists of a shared EfficientNet \cite{tan2019efficientnet} backbone with a two-way Feature Pyramid Network (FPN) and separate semantic and instance segmentation heads.
The semantic segmentation head consists of modified Dense Prediction Cell (DPC) \cite{chen2018searching} modules to capture the contextual information of features at multiple scales.
The instance head is a variation of Mask RCNN \cite{he2017mask}.
As EfficientPS provides only panoptic segmentation, its segmentation heads lack the capability to estimate the uncertainty of its predictions.
We employ evidential deep learning \cite{sensoy2018evidential} to quantify semantic segmentation, instance segmentation, and classification uncertainty in our uncertainty-aware semantic and instance segmentation heads.

\subsubsection{Uncertainty-aware semantic head}

We modify the semantic head output by replacing the softmax layer with the softplus activation function. The output of the softplus activation provides the evidence signal, which is the measure of the amount of support collected from the data to assign the pixel to a particular class.
The authors \cite{sensoy2018evidential} proposed ReLU as activation function, but we found that softplus performs better, as presented in the supplementary material\footnote[1]{Supplementary material can be found at: \url{https://arxiv.org/abs/2206.14554}} (Sec. S.1.2) . 

We use the Dirichlet distribution as prior for our per-pixel multinomial classification, parametrized by $\alpha = [\alpha^{1}, ..., \alpha^{C}]$,
where $C$ is the number of classes and
$\alpha^c_i = \text{softplus}(l^c_i) + 1$
for network output logit $l$ for pixel $i$ and class $c$.
The corresponding probability $p$ and uncertainty $u$ are calculated as:
\begin{align}
p^c_i &= \alpha^c_i / S_i \label{prob_equation} \\
u_i &= C / S_i\label{unc_equation},
\end{align}
where $S_i = \sum_{c=1}^C  \alpha^c_i$. Note that the uncertainty  $u_i$ is inversely proportional to $S_i$ as $C$ is constant.

To train our semantic head we use the type-II maximum likelihood version of the loss
and modify it to suit the semantic segmentation setting:
 \begin{equation}\label{eq:log_loss_sem}
\mathcal{L}_{\log}^\text{sem} = \sum_{i=1}^N\sum_{c=1}^C y^c_i \log(S_i / \alpha^c_i).
\end{equation}
Here, $N = W \times H$ is the number of pixels, and $y$ is the one-hot encoded vector with $y^c=1$ if $c$ is the ground truth class, else $y^c=0$.
We found the proposed mean square error loss (MSE) \cite{sensoy2018evidential} not suited for our multi-class segmentation task, as presented in the supplementary material (Sec. S.1.1).

We further employ the the Lov\'{a}sz softmax loss of Berman \emph{et al.} \cite{berman2018lovasz}
for training our semantic head.
However, we modify it by replacing the softmax with the evidential probabilities of \eqref{prob_equation},
and call it Lov\'{a}sz Evidential loss $\mathcal{L}_\text{LE}$.

Similar to \cite{sensoy2018evidential} we add a KL-term to force the evidence created by non-ground truth logits to a minimum, such that the network predict strong evidence only for the correct class, suppressing the others. The overall loss for the semantic head is
\begin{equation}\label{eq:log_loss_kl_sem}
\mathcal{L}_\text{sem} = \mathcal{L}_{\log}^\text{sem} + \lambda_t \mathcal{L}_\text{KL}^\text{sem} + \mathcal{L}_\text{LE},
\end{equation}
where $\lambda_t$ controls the weight of the KL term.
We use $\lambda_t = 0.06 \times \text{min}\{1, t/(60I)\}$, where $t$ is the current training iteration and $I$ is the number of iterations per epoch.
Thus $\lambda_t$ increases until $60$ epochs and then stays at the value of $0.06$. 
This way network first learns to segment correctly and later also focuses on
providing correct uncertainty \cite{sensoy2018evidential}.
We provide the supporting study regarding our choice in \secref{sec:lambda}.

\subsubsection{Uncertainty-aware instance head}

The instance head is a modified version of Mask RCNN \cite{he2017mask} similar to the EfficientPS architecture. The head consists of a Region Proposal Network (RPN) to generate proposals and objectness scores.
The ROI align extracts the features bounded within the generated proposals from the RPN. These features are fed to the separate bounding box regression, classification, and mask generation heads. We focus on providing the uncertainty-aware mask segmentation and object classification.

The mask head creates a $K \times W \times H$ output for each bounding box, where $K$ is the number of thing classes and $W = H = 28$. We replace the sigmoid in the original implementation with softplus and utilize the same loss function as for the semantic head (\eqref{eq:log_loss_sem}) to calculate $\mathcal{L}_\text{log}^\text{mask}$.
Together with the KL regularization for false evidence, the mask loss is
\begin{align}
\mathcal{L}_\text{mask} = \mathcal{L}_\text{log}^\text{mask} + \lambda_{\text{t}} \mathcal{L}_\text{KL}^\text{mask}.
\end{align}

For the classification we directly use the loss together with the KL term provided by \cite{sensoy2018evidential}:
 \begin{align}
\mathcal{L}_\text{cls} = \sum_{k=1}^K y^k \log(S / \alpha^k) + \lambda_{\text{t}} \mathcal{L}_\text{KL}^\text{cls}.
\end{align}

The classification and mask prediction probabilities and uncertainty are obtained using \eqref{prob_equation} and \eqref{unc_equation}, respectively.
The losses for bounding box regression (bbox), the RPN objectness score (os) and object proposal (op) are similar to EfficientPS.
This yields an overall training loss for the instance segmentation of
\begin{align}
    \mathcal{L}_\text{inst} = \mathcal{L}_\text{mask} + \mathcal{L}_\text{cls} + \mathcal{L}_\text{bbx} + \mathcal{L}_\text{os} + \mathcal{L}_\text{op}.
\end{align}

Finally, our overall training loss is
\begin{equation}\label{eq:overallloss}
\mathcal{L} = \mathcal{L}_\text{sem} + \mathcal{L}_\text{inst}.
\end{equation}

\subsection{Improved Panoptic Fusion Module}\label{fusion}

\begin{figure*}
\centering
\includegraphics[width=0.9\textwidth]{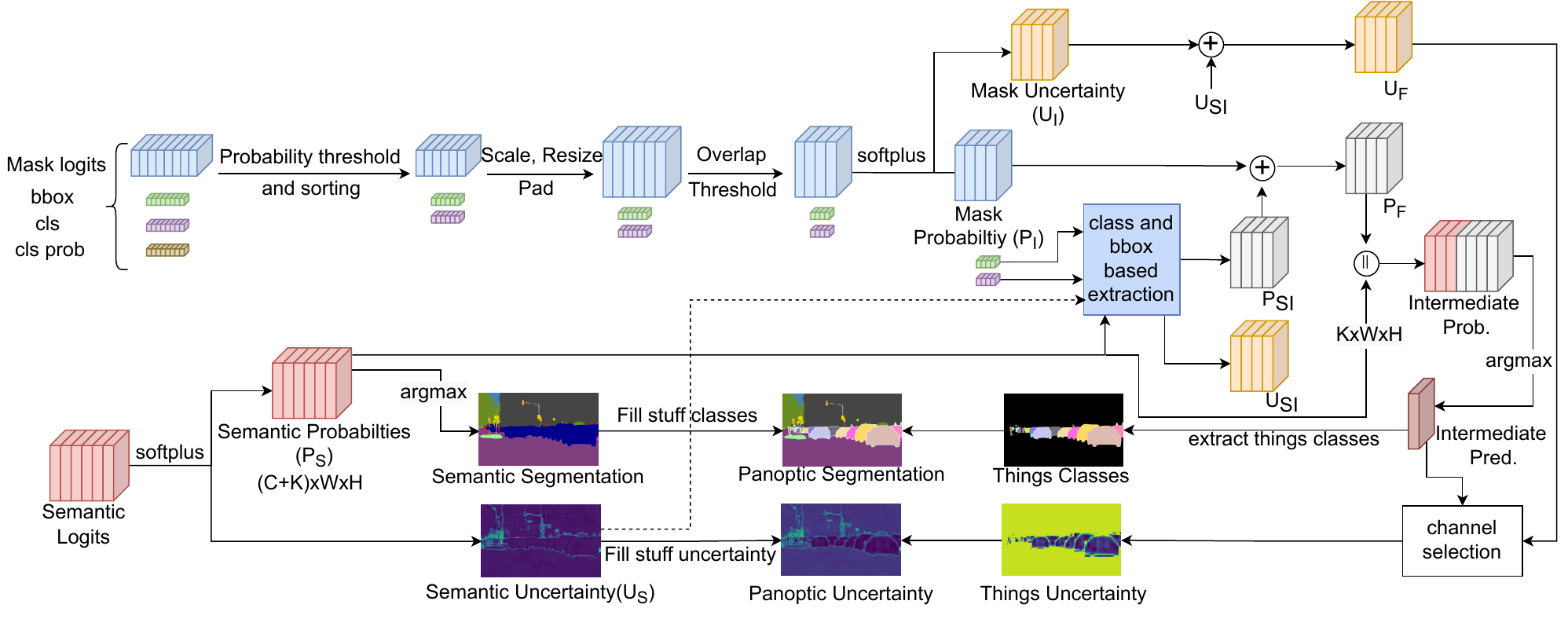}
\caption{Illustration of our proposed panoptic fusion module. A description is given in the text.
} 
\label{fig:fusionmodule}
\end{figure*}

The fusion of the semantic and instance segmentations is an intricate task due to their inherent overlap and scale difference.
To solve this task, we propose a novel fusion module, illustrated in \figref{fig:fusionmodule},
which leverages the predicted probabilities and uncertainties of our semantic and instance segmentation heads.

First, we obtain a set of instances from the instance segmentation head with their corresponding mask logits, bounding box, class, and class probabilities.
The latter are calculated using \eqref{prob_equation} and are then used to discard instance predictions below a probability of $0.5$.
Next, we resize the mask logits to match the input image resolution.
To solve the overlap problem among instance masks, we first take the sigmoid of the mask logits and convert them into binary masks.
Then, if the overlap between the masks is greater than 0.5, we keep the one with the highest probability and discard the others.
After a softplus operation, we apply \eqref{prob_equation} and (\ref{unc_equation}) to convert the logits into instance mask probability and uncertainty tensors P$_\text{I}$ and U$_\text{I}$, respectively.
\rule{0pt}{0.5ex}    

Similarly, we employ the semantic segmentation logits to obtain corresponding probability and uncertainty tensors P$_\text{S}$ and U$_\text{S}$, respectively.
These are converted into instance mask probability and uncertainty tensors P$_\text{SI}$ and U$_\text{SI}$, respectively,
by only keeping the values from P$_\text{S}$ and U$_\text{S}$ within the bounding box provided by a particular instance and zeroing out all other values.
To fuse these with the corresponding primary tensors, it suffices to take the averages: $\text{P}_\text{F} = (\text{P}_\text{I} + \text{P}_\text{SI})/2$ and
uncertainties $\text{U}_\text{F} = (\text{U}_\text{I} + \text{U}_\text{SI})/2$.
This averaging allows to consider the semantic uncertainty within each instance bounding box as well.
Note that we do not have to deal with scaling issues here, since we are fusing probabilities and uncertainties instead of logits.
The magnitude of the final uncertainty is determined by calibration, as discussed in Section \ref{sec:lambda}.

We concatenate the semantic segmentation probabilities P$_\text{S}$ and fused probabilities P$_\text{F}$ and apply argmax to obtain an intermediate panoptic prediction.
Similarly, we concatenate U$_\text{S}$ and U$_\text{F}$ and obtain the corresponding intermediate panoptic uncertainties by selecting the channel provided by the previous argmax operation.
Finally, we extract the \textit{thing} predictions from intermediate panoptic result and fill the rest with \textit{stuff} predictions to obtain final panoptic segmentation and uncertainties.

\subsection{Uncertainty-aware Panoptic Segmentation Task}

The goal of uncertainty-aware panoptic segmentation is to augment panoptic segmentation by an estimate of the underlying prediction uncertainty. Accordingly, given $C$ semantic classes, our uncertainty-aware panoptic segmentation maps each pixel $i$ in the image to a set $(c_i, k_i, u_i) \in C \times \mathbb{N} \times U$, with $U = [0,1]$.
Here, $c_i$ signifies the semantic class, $k_i$ is the unique id that groups the pixels belonging to the same semantic class but to separate instances, and $u_i$ represents the per-pixel uncertainty associated with the prediction. 
In the case of a \textit{stuff} class, $u_i$ represents the predicted uncertainty of the pixel belonging to the semantic class $c_i$. 
For a \textit{thing} class instance, $u_i$ defines the network's uncertainty in predicting a pixel belonging to class $c_i$ and instance $k_i$.

A desired outcome is that large uncertainties will be predicted for pixels that are prone to either be wrongly classified or assigned to a wrong instance.
Typically, these can be pixels of unknown objects, pixels at the border of objects, or even pixels in out-of-distribution images.
We will evaluate this behavior in Section~\ref{sec:experiments}.

\subsection{Metrics for uncertainty-aware panoptic segmentation}

In this section we introduce two novel metrics designed for evaluating uncertainty-aware panoptic segmentation.
These metrics allow to judge the quality of the network prediction, i.e. the pixel-wise class and instance outputs, in combination with the predicted uncertainty.

\paragraph{Calibration metric}
Here, we aim to provide a metric capable of evaluating the network calibration, i.e. how well the predicted confidence matches the actual accuracy the prediction, for the panoptic segmentation task.
A common measure for the calibration accuracy is the Expected Calibration Error (ECE) \cite{naeini2015obtaining}.
However, as pointed out by Nixon \emph{et al.} \cite{nixon2019measuring}, the ECE has some severe limitations, in particular: 
First, it takes only the maximum class probability of the prediction into account, ignoring the probability of all other classes. 
Second, it is only suited to evaluate the calibration of the semantic segmentation task, while the instance segmentation is ignored.

To solve the first limitation, we propose to employ the predicted uncertainty, rather than just the highest class probability.
Considering $u_i \in [0,1]$ to be the predicted uncertainty for pixel $i$, we define the corresponding confidence as $\text{conf}_i = 1 - u_i$.
Further, we define the corresponding accuracy as $\text{acc}_i = 1$
if the predicted class matches the ground truth, and $\text{acc}_i = 0$ otherwise.
For each image we partition $\text{conf}_i$ into $M$ equally spaced bins and calculate the average confidence $\text{conf}(B_m)$ and average accuracy $\text{acc}(B_m)$ as in \cite{guo2017calibration} for each bin $B_m$.
Then we define a uncertainty-aware calibration metric as
 \begin{align}
\text{uECE} = \sum_{m=1}^{M} \frac{|B_m|}{N} \left| \text{acc}(B_m) - \text{conf}(B_m) \right|,
\end{align}
where $|B_m|$ is the number of pixels in bin $B_m$, and $N$ is the total number of pixels.
This definition is analogous to the original ECE, but using the confidence instead of the highest probability output, solving the first limitation. 

To solve the second limitation, we proceed to additionally incorporate the instances of the scene into a metric.
First, we identify correctly predicted segments $f$ as those that have $\text{IoU} > 0.5$ with a ground truth segment $g$ of the same class.
Here, a segment can either be the mask of a single instance of a \textit{thing} class, or all pixels belonging to a \textit{stuff} class.
We calculate uECE$(f,g)$ separately for each of these unique matching pairs $(f,g)$ \cite{kirillov2019panoptic}.
Here, all pixels of $f$ are taken into account, however, the pixels that are not part of $g$ receive $\text{acc}_i = 0$. 
Further, we calculate uECE$(\tilde{f})$ for each of the remaining (unmatched) predicted segments, $\tilde{f}$, separately, by setting $\text{acc}_i = 0$ for all pixels.
Note that $f$ are true positives and $\tilde{f}$ are false positives.
We then take the average over all $M$ predicted segments to form our novel panoptic calibration metric:
\begin{align}
\text{pECE} = \frac{1}{M} \left( \sum_{(f,g)} \text{uECE}(f, g) +  \sum_{\tilde{f}} \text{uECE}(\tilde{f}) \right).
\end{align}
This quantity is sensitive to the calibration within the instances of \textit{thing} classes, as well as to the \textit{stuff} classes, taking the boundaries of all these segments into account.
False negatives are taken into account as well, since panoptic segmentation guarantees a prediction for every pixel, and thus, false negatives will either be a part of the matched or the unmatched segments.

\paragraph{Overall performance metric}
Our next aim is to provide a unified metric to evaluate the panoptic segmentation and uncertainty prediction together,
trading off between segmentation performance and calibration accuracy.
We base it on the common Panoptic Quality (PQ): 
\begin{align}
\text{PQ} = \frac{\sum_{(f,g)} \text{IoU}(f,g)}{\text{TP} + \frac{1}{2}\text{FP} + \frac{1}{2}\text{FN}} \,
\end{align}
where the IoU is calculated for the matched pairs $(f,g)$, and TP, FP and FN are the number of true positive, false positive, and false negative segments, respectively.
We define our novel uncertainty-aware panoptic quality uPQ by combining pECE with PQ as:
\begin{align}
\text{uPQ} = (1-\text{pECE}) \text{PQ}
\end{align}

Thus, if the network is perfectly calibrated, pECE would be close to zero, and the overall uPQ will signify the PQ.
On the other hand, if the network is poorly calibrated, it will be reflected in a lower uPQ.
In this way, we can consider both the quality of panoptic segmentation prediction and the calibration of predicted uncertainty.
Similar to PQ, uPQ can be calculated separately for each class.

\section{Experimental Evaluation}\label{sec:experiments}

\subsection{Datasets}

We evaluate the performance of our approach on two datasets.
First, the Cityscapes dataset \cite{Cordts2016Cityscapes}, consisting of 5000 densely annotated images from urban scenarios with 11 \textit{stuff} and 8 \textit{thing} class objects.
By default the dataset is split into default-training (2975 images), default-validation (500 images), and default-testing (1525 images).
However, there are no annotations available for the default-testing images.
Hence, we created our own split by using the default-validation set as our testing set (for evaluating the final results), and
by dividing the default-training set into our training set with 2562 images and our validation set with 413 images (for performance tracking and parameter tuning).
All results presented in this paper are calculated from our testing (= default-validation) set unless stated otherwise.

For testing the  performance of the network on out-of-distribution samples, we choose the challenging Indian Driving dataset \cite{varma2019idd}.
It consists of far more instances of \textit{thing} classes than other datasets and often lacks structural cues like lanes and sidewalks.
Similar to above, we use the default-validation set to evaluate our method. No training or tuning was done with the Indian Driving dataset.

\subsection{Baselines}

We provide five baselines for our proposed task of uncertainty-aware panoptic segmentation on the Cityscapes dataset.
The first two baselines are state-of-the-art top-down and bottom-up networks, EfficientPS and PanopticDeepLab \cite{cheng2020panoptic}, for the panoptic segmentation task without uncertainty-awareness.

As we focus on autonomous driving, we need sampling-free methods for the real-time uncertainty estimation. We choose temperature scaling (TS) and evidential learning (Ev) for this purpose.
For TS we train the network on the training set, freeze the whole network, and learn a logit scaling parameter (temperature) for the semantic segmentation head. This parameter is learned on our validation set, minimizing the cross-entropy loss.
This yields our next two baselines, EfficientPS+TS and PanopticDeepLab+TS.

We estimate the uncertainty output for the original networks by calculating the normalized entropy of the predicted probabilities.
However, for the TS variants, we utilize maximum probability as confidence value for calculating uECE, since the temperature scaling aims to calibrate this quantity.
The final baseline PanopticDeepLab+EV includes replacing the softmax with the softplus operation and training the semantic segmentation head with the evidential loss
together with the KL term of \eqref{eq:log_loss_kl_sem}.
We train the baselines on a smaller training set and with smaller batch sizes than the original papers leading to minor performance drops. 
Specifically, we utilize a batch size of 8 for EfficientPS \cite{mohan2021efficientps} and 24 for PanopticDeepLab \cite{cheng2020panoptic}, except the original batch sizes of 16 and 32, respectively.
We compare these baselines to our EvPSNet and to a variant of it, EvPSNet*, which was trained without Lov{\'a}sz Evidential loss.

\subsection{Training Procedure}

We train EvPSNet using the image size of 2048x1024 pixels for the Cityscapes dataset. We augment the data by performing flipping and scaling within the range of [0.5,2.0]. We initialize the backbone of our network with weights from the EfficientNet model trained on the ImageNet dataset and Xavier initialization for the other layer with zero bias. We optimize the network using stochastic gradient descent with a momentum of 0.9 for 160 epochs with a multi-step learning rate schedule. We use a batch size of 8 and an initial learning rate of 0.07 and decay it twice by a factor of 0.1 on the epochs 120 and 144.

\begin{table*}
\begin{center}
\rule{0pt}{0.5ex}    
\caption{Performance values in \% on the Cityscapes data. Lower values are better for $\downarrow$, and larger values otherwise.}
\label{tab:cityval}
\footnotesize
\begin{tabular}
{l|>{\color{tablecolor}}c>{\color{tablecolor}}c>{\color{tablecolor}}c>{\color{tablecolor}}c|>{\color{tablecolor}}c>{\color{tablecolor}}c>{\color{tablecolor}}c>{\color{tablecolor}}c|>{\color{tablecolor}}c>{\color{tablecolor}}c>{\color{tablecolor}}c>{\color{tablecolor}}c|>{\color{tablecolor}}c}
\toprule
Method & uPQ & PQ  & SQ & RQ & uPQ\textsuperscript{Th}& PQ\textsuperscript{Th} & SQ\textsuperscript{Th} & RQ\textsuperscript{Th} & uPQ\textsuperscript{St} & PQ\textsuperscript{St} & SQ\textsuperscript{St} & RQ\textsuperscript{St} &pECE $\downarrow$ \\
\midrule
EfficientPS \cite{mohan2021efficientps} & $46.7$ &$63.0$ &  $81.4$ & $76.3$& $40.8$ & $\mathbf{57.5}$ & $80.3$ & $\mathbf{71.4}$& $51.3$ & $67.0$ & $82.3$ & $79.9$ & $25.9$ \\
EfficientPS \cite{mohan2021efficientps} + TS \cite{guo2017calibration} & $47.7$ &$62.9$ &  $81.4$ & $76.3$& $41.0$ & $57.3$ & $80.2$ & $71.3$& $52.9$ & $67.0$ & $82.3$ & $79.9$ & $24.2$ \\
PDeepLab \cite{cheng2020panoptic} & $39.9$& $61.9$ &  $81.9$ & $74.5$& $32.6$ & $52.8$ & $80.6$ & $65.1$& $45.5$ & $68.5$ & $82.9$ & $81.2$ & $35.5$ \\
PDeepLab \cite{cheng2020panoptic} + TS \cite{guo2017calibration} &$46.6$ &$61.9$ & $81.9$ & $74.5$& $37.2$ & $52.8$ & $80.6$ & $65.1$& $53.9$ & $68.5$ & $82.9$ & $81.2$ & $24.7$ \\
PDeepLab \cite{cheng2020panoptic} + Ev &$46.6$ &$61.4$ &  $81.9$ & $74.0$& $36.4$ & $51.8$ & $\mathbf{80.8}$ & $64.2$& $54.7$ & $68.4$ & $82.8$ & $81.2$ & $24.1$ \\
\midrule
EvPSNet w/o LE & $49.1$ & $62.3$ & $81.0$ & $75.9$ & $41.1$& $53.9$ & $79.8$ & $67.3$ & $55.2$& $68.4$ & $81.8$ & $82.2$ & 21.2 \\
EvPSNet (ours) & $\mathbf{51.5}$ & $\mathbf{63.7}$ & $81.3$ & $\mathbf{77.5}$ & $\mathbf{45.6}$& $56.4$ & $80.2$ & $70.3$ & $\mathbf{55.7}$& $\mathbf{69.0}$ & 82.2 & $\mathbf{82.7}$ & $\mathbf{19.3}$\\
\bottomrule
\end{tabular}
\end{center}
\end{table*}

\subsection{Quantitative Results}

In this section, we report results to quantitatively compare the performance of our proposed EvPSNet architecture against the introduced baselines. To this end, we employ our proposed uncertainty-aware panoptic segmentation metrics, as well as standard panoptic segmentation metrics. 

The results for the Cityscapes data are presented in \tabref{tab:cityval}.
Our proposed EvPSNet model outperforms all the proposed baselines on the introduced uncertainty-aware panoptic segmentation metrics uPQ and pECE by a margin of 3.8\% and 
\myworries{4.9\%} respectively, in comparison to EfficientPS+TS, the second-best baseline. Also, note that PanopticDeepLab+Ev uses our formulation of \eqref{eq:log_loss_sem} together with the softplus activation function.
In comparison to EfficientPS, our network achieves a gain of \myworries{4.8\%} in uPQ, \myworries{4.8\%} in uPQ$^{\text{Th}}$, \myworries{4.4\%} in uPQ$^{\text{St}}$ and \myworries{6.6\%} in pECE.
With the advantage of our proposed Lov{\'a}sz evidential loss function for the semantic segmentation, our network achieves the highest value of \myworries{69.0\%} for PQ$^{\text{St}}$.  Moreover, we also achieve the highest PQ value, which we attribute to better uncertainty and probability estimation, which is further utilized by our proposed panoptic fusion module. PQ$^{\text{Th}}$ of EvPSNet decreases in comparison to EfficientPS, which is due to only utilizing the segmentation uncertainties for the instance segmentation head. We believe that incorporating the regression uncertainties for the bounding box can improve results. Nevertheless, EvPSNet gains \myworries{4.8\%} on the uPQ$^{\text{Th}}$ in comparison to EfficientPS. 

While the addition of temperature scaling does not significantly affect the performance of the base networks on the classical panoptic metrics,
it does improve the uncertainty estimation (pECE and thus uPQ).
However, the improvement is not as large as from our evidential method.

Additionally, we show the advantage of using the Lov{\'a}sz Evidential (LE) loss function by comparing our EvPSNet (with LE) with a version trained without it. Our modified loss function gains \myworries{1.4\%} on uPQ and 0.7\% on PQ.

\begin{figure*}
\centering
\footnotesize
\setlength{\tabcolsep}{0.05cm}
{
\renewcommand{\arraystretch}{0.2}
\newcolumntype{M}[1]{>{\centering\arraybackslash}m{#1}}
\begin{tabular}{cM{0.2\linewidth}M{0.2\linewidth}M{0.2\linewidth}M{0.2\linewidth}}
& Input Image & Panoptic Segmentation & Uncertainty Map & Error Map \\
\\
\\
\\
\rotatebox[origin=c]{90}{(a)}& {\includegraphics[width=\linewidth, frame]{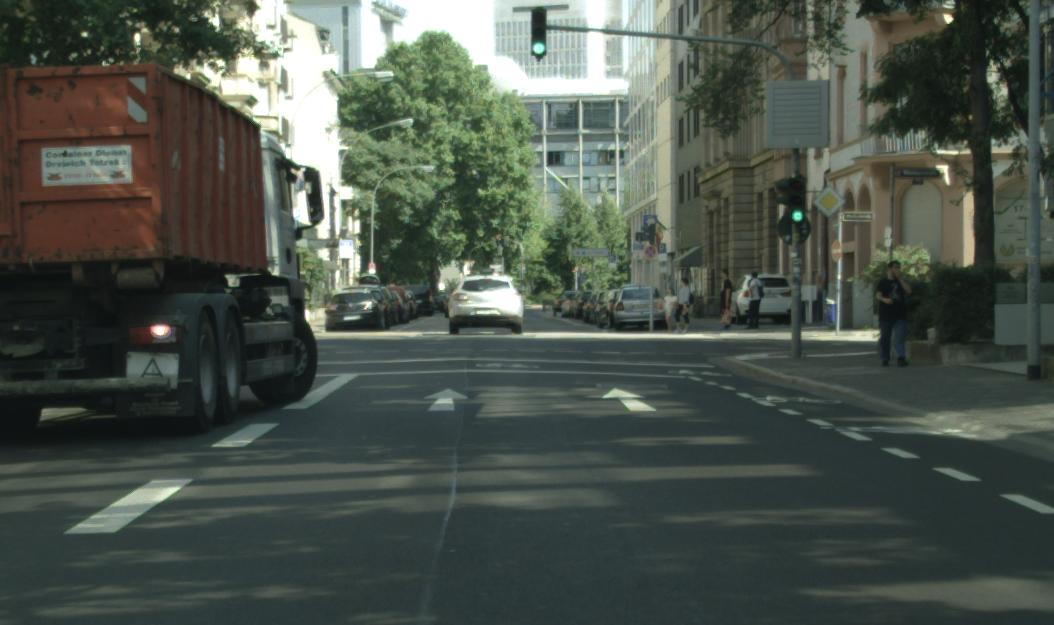}} & {\includegraphics[width=\linewidth, frame]{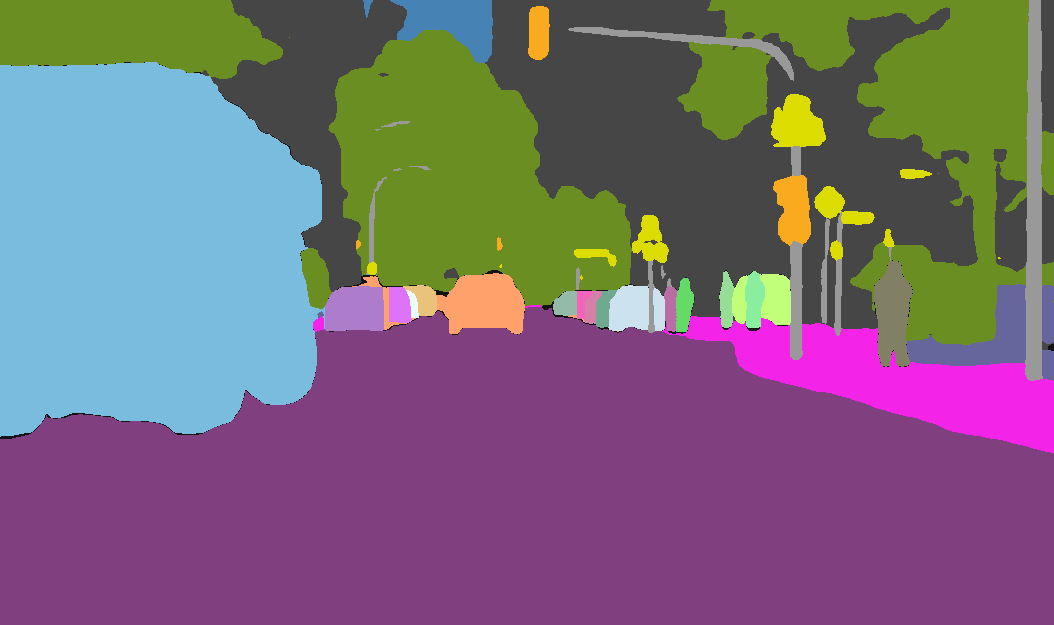}} & {\includegraphics[width=\linewidth, frame]{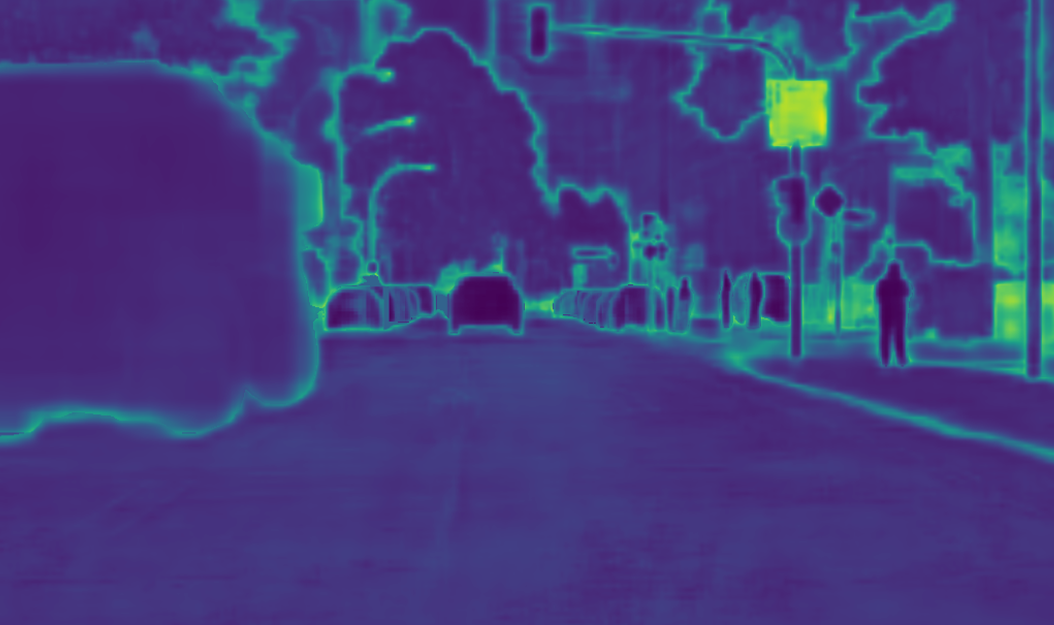}} & {\includegraphics[width=\linewidth, frame]{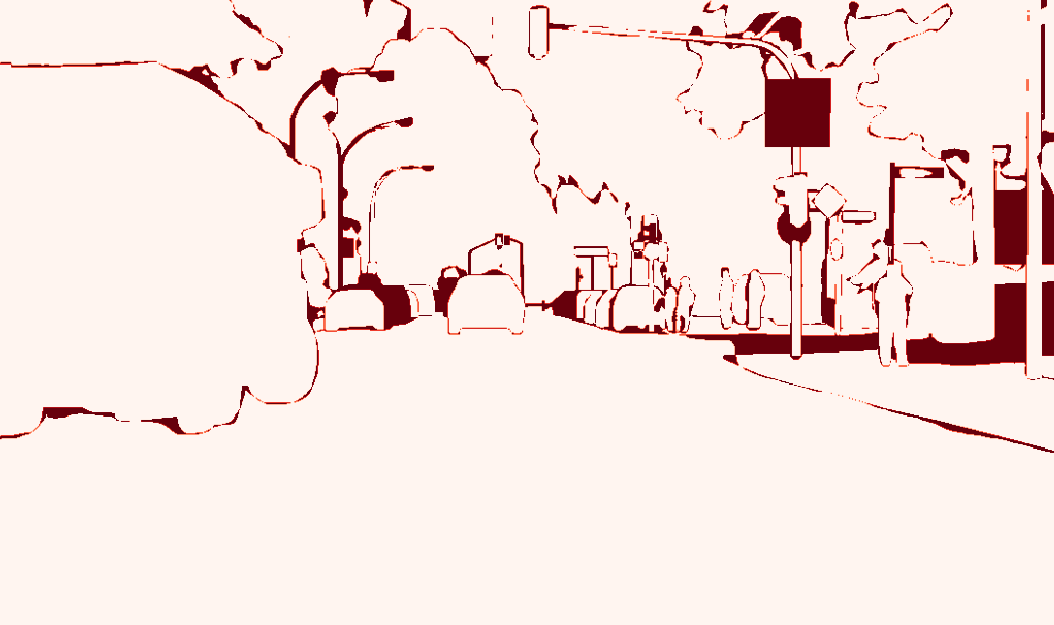}} \\
\\
\rotatebox[origin=c]{90}{(b)}& {\includegraphics[width=\linewidth, frame]{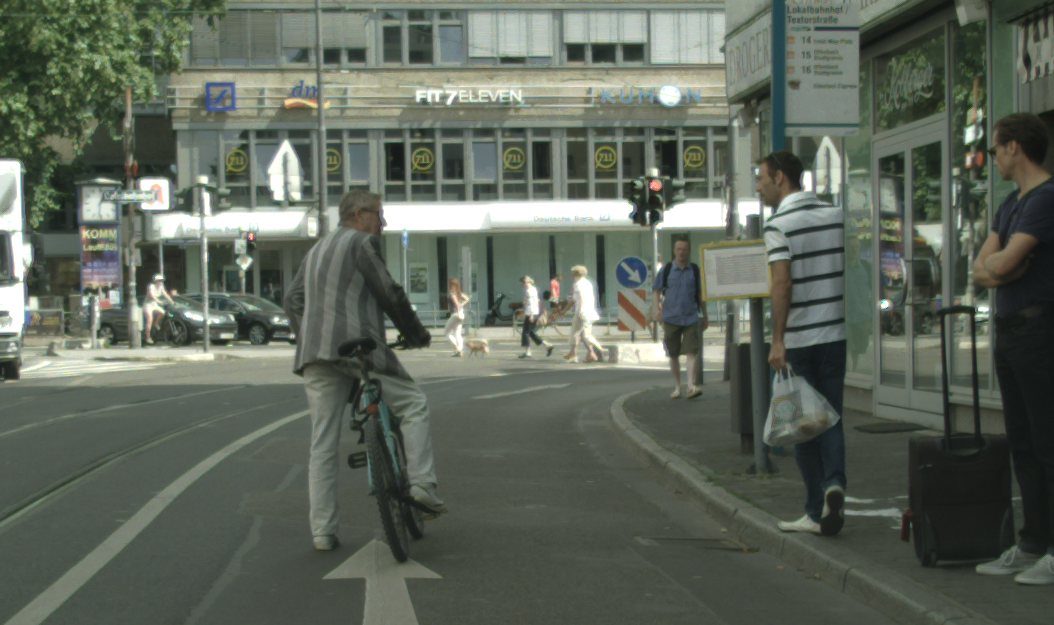}} & {\includegraphics[width=\linewidth, frame]{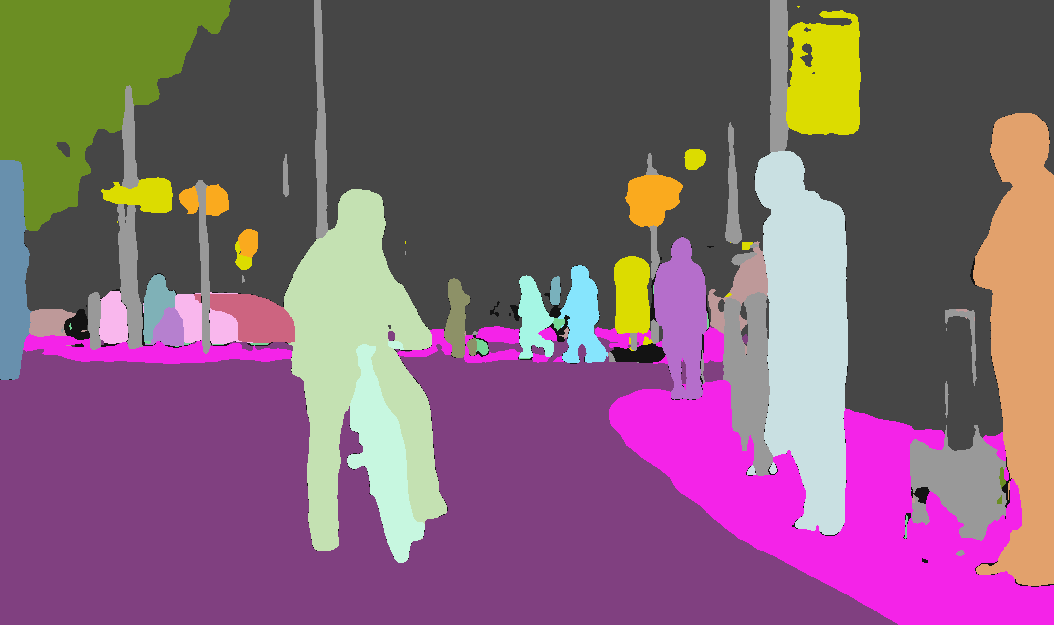}} & {\includegraphics[width=\linewidth, frame]{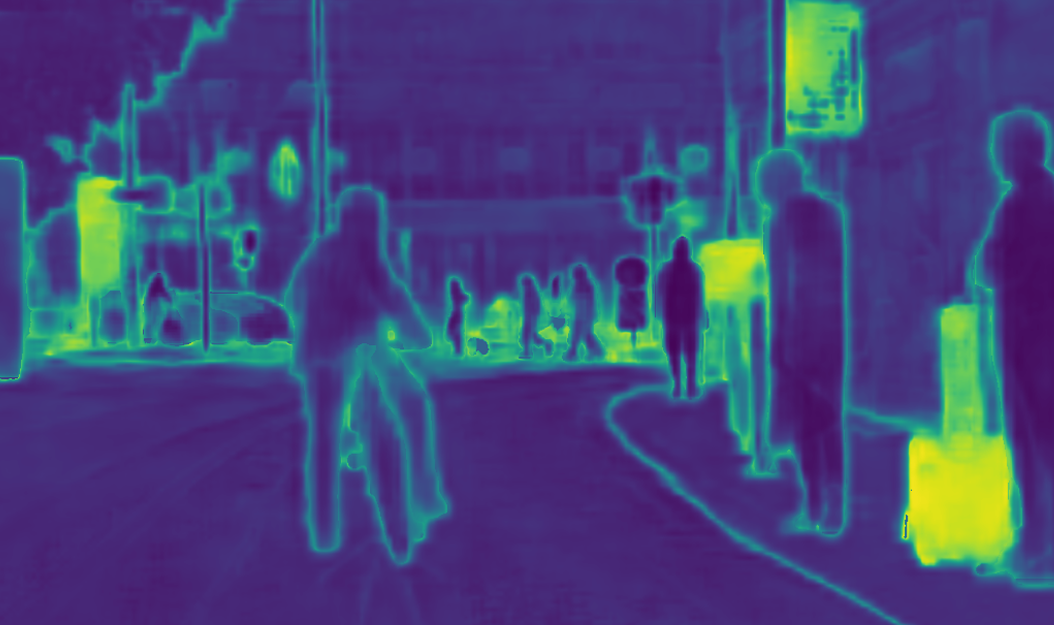}} & {\includegraphics[width=\linewidth, frame]{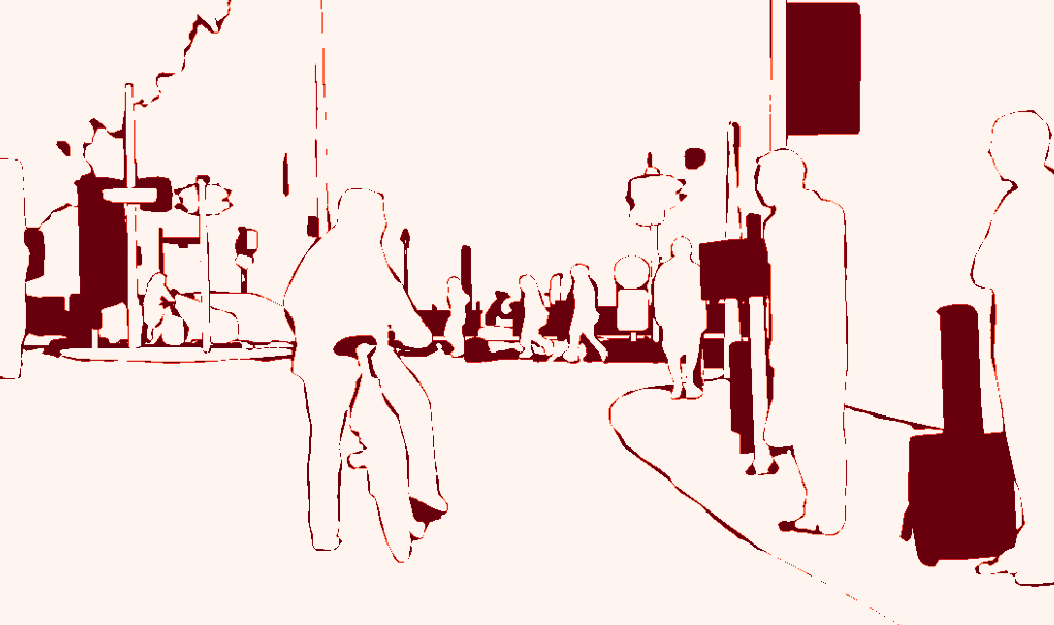}} \\

\end{tabular}
}
\caption{Qualitative results for the uncertainty-aware panoptic segmentation by our EvPSNet for two images of the Cityscapes data.
Bright regions in the uncertainty map depict high predicted uncertainty, and dark regions in the error map depict misclassified pixels.}
\label{fig:qualitative results}
\vspace{-0.4cm}
\end{figure*}

\subsection{Qualitative Results}

We provide qualitative results of our EvPSNet, including the panoptic segmentation, uncertainty and error maps in \figref{fig:qualitative results}.
Comparing the predicted uncertainty with the error maps a high correlation can be observed,
which provides clear visual evidence of the validity of the former.
For example, the network is not able to correctly detect some signs, street lights and a suitcase (dark regions in the error map), which is clearly correlated with higher predicted uncertainties.
Another source of error are the outlines objects, clearly reflected in the uncertainties as well.
Minor failure cases can be found, e.g. in \figref{fig:qualitative results}(a) a traffic sign was falsely predicted as part of an overlapping traffic light (orange) with relatively low uncertainty.

\subsection{Ablation studies}
In this section, we provide quantitative and qualitative analysis of our design choices for different components of our proposed network and metrics.

\subsubsection{Calibration metric}\label{sec:eval_metric}

This study compares the original ECE metric, utilizing the maximum probability, to the proposed uncertainty-based uECE and the panoptic pECE.
From \tabref{tab:ECE_met} it is evident that ECE and uECE are correlated, with different tendencies in particular for smaller values.
Differences are expected, since ECE only utilizes the maximum probability output, while uECE utilizes the overall uncertainty estimate of the network.

The overall magnitude of the pECE is higher than the other two, which is expected given that it
gives more weight to the instances and their boundaries, which are typically the most uncertain parts of the prediction.
Considering only the estimation of semantic segmentation uncertainties is an easier task, since the network can assign high confidences to all the grouped instances in close vicinity, as shown in \figref{fig:sem_unc}.
This is also evident for PanopticDeepLab+Ev, which has a low uECE, but a high pECE. We attribute this to the lack of instance-aware uncertainty estimation.

\begin{table}
\begin{center}
    \caption{Various ECE values in \% on the Cityscapes data. Lower values are better.}
\label{tab:ECE_met}
\footnotesize
\begin{tabular}
{l|>{\color{tablecolor}}c|>{\color{tablecolor}}c|>{\color{tablecolor}}c}
\toprule
Method & ECE & uECE & pECE \\
\midrule
EfficientPS \cite{mohan2021efficientps} & 12.3 & 17.6 & 25.9 \\
EfficientPS \cite{mohan2021efficientps} + TS \cite{guo2017calibration} & $\phantom{0}\mathbf{4.5}$ & \phantom{0}5.2& 24.2 \\
PDeepLab \cite{cheng2020panoptic} & $17.4$ &$26.7$&$35.5$\\
PDeepLab \cite{cheng2020panoptic} + TS \cite{guo2017calibration} & \phantom{0}6.0 & \phantom{0}7.1& 24.7 \\
PDeepLab \cite{cheng2020panoptic} + Ev & \phantom{0}5.6& $\phantom{0}\mathbf{3.0}$ & $24.1$\\
\midrule
EvPSNet (ours) & \phantom{0}5.3 & \phantom{0}3.5 & $\mathbf{19.3}$ \\
\bottomrule
\end{tabular}
\end{center}
\end{table}

\begin{table}
\begin{center}
\caption{Fusion module performance values in \% on Cityscapes data. Higher values are better.}
\label{tab:FUSION}
\footnotesize
\begin{tabular}
{l|>{\color{tablecolor}}c>{\color{tablecolor}}c>{\color{tablecolor}}c|>{\color{tablecolor}}c|>{\color{tablecolor}}c}
\toprule
Method &  PQ  & SQ & RQ & PQ\textsuperscript{Th} & PQ\textsuperscript{St}\\
\midrule
Seamless \cite{porzi2019seamless}&$59.5$ & $77.4$& $75.2$ &$47.6$ &$68.1$ \\
EfficientPS \cite{mohan2021efficientps}&$62.4$& $80.7$& $76.2$ &$54.8$&$67.9$\\
\midrule
Ours & $\mathbf{63.7}$ & $\mathbf{81.3}$ & $\mathbf{77.5}$  & $\mathbf{56.4}$ & $\mathbf{69.0}$ \\
\bottomrule
\end{tabular}
\end{center}
\vspace{-4mm}
\end{table}
\subsubsection{Panoptic fusion}

We compare the performance of our proposed panoptic fusion module with the ones proposed by Seamless \cite{porzi2019seamless} and EfficientPS \cite{mohan2021efficientps}.
We train a network based on the our evidential settings for this experiment and replace only the panoptic fusion module.
All methods utilize the same instance and semantic segmentation outputs.
Seamless fusion is done by simply taking instance segmentation results for all \textit{thing} classes and semantic segmentation for \textit{stuff} class.
Hence, this method solves the inherent overlap problem by discarding possibly valuable information.
EfficientPS combines the logits from both segmentations in a deterministic fashion and then calculates the panoptic output.

Since the baseline fusion methods do not provide any confidence or uncertainty estimate, we evaluate only the panoptic metrics, listed in \tabref{tab:FUSION}.
Utilizing our improved fusion leads to an improvement of \myworries{1.1\%} for PQ$^{\text{St}}$, \myworries{1.6\%} for PQ$^{\text{Th}}$, and \myworries{1.3\%} for the overall PQ in comparison to the EfficientPS fusion.
We attribute the performance gain to utilizing the probabilities in the fusion, which solves the overlap problem straight forward, yielding a better combination of the different segmentations.

\begin{table}
\begin{center}
    \caption{Various ECE values in \% on the Indian Driving data. Lower values are better.}
\label{tab:IDD_ECE_met}
\footnotesize
\begin{tabular}
{l|c|c|c}
\toprule
Method & ECE & uECE & pECE  \\
\midrule
EfficientPS \cite{mohan2021efficientps} & 19.7 & 20.3 & 40.0 \\
EfficientPS \cite{mohan2021efficientps} + TS \cite{guo2017calibration} & 11.5 & \phantom{0}9.3& 35.7 \\
\midrule
EvPSNet (ours) & $\mathbf{\phantom{0}5.7}$ & $\mathbf{\phantom{0}7.5}$ & $\mathbf{33.0}$ \\
\bottomrule
\end{tabular}
\end{center}
\vspace{-4mm}
\end{table}
\subsubsection{Effect of $\lambda_{\text{t}}$} \label{sec:lambda}
\begin{figure}
\centering
\includegraphics[width=0.9\linewidth]{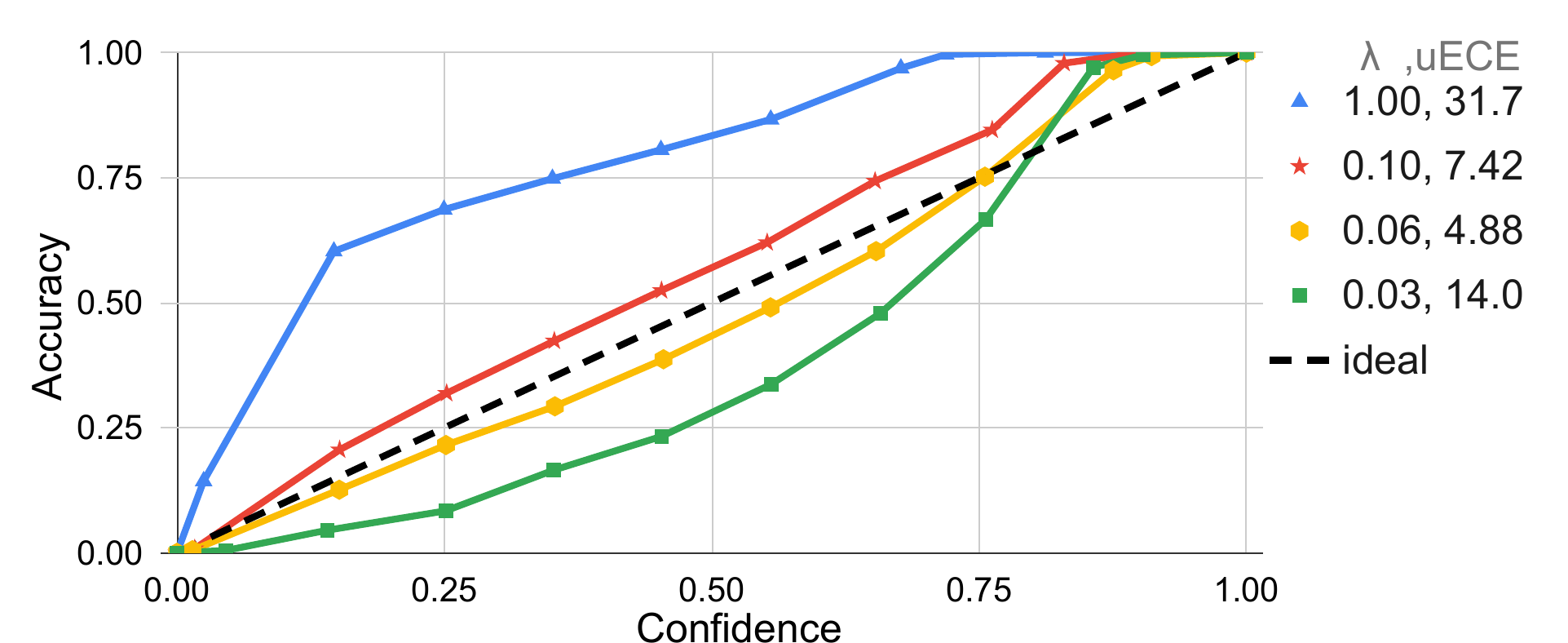}
\caption{Calibration curves for various $\lambda_{\text{t}}$, and resulting uECE [\%].} 
\label{fig:lambda}
\end{figure}

The authors of evidential deep learning suggest using a KL term in the loss function to decrease the evidence provided by confusing samples \cite{sensoy2018evidential}.
Furthermore, they suggest gradually increasing the weight, $\lambda_{\text{t}}$, of the KL term until it reaches a value of 1.0 and then keeping this for the rest of the training.
In our experiments, we found that this setting leads to underconfident results and tried different final values of $\lambda_{\text{t}}$.
The resulting calibration curves for our validation set are shown in \figref{fig:lambda}.
It is apparent that this parameter controls the confidence of the network.
Finally, we used a value of $\lambda_{\text{t}} = 0.06$, which yielded the best uECE in our case.
\subsubsection{Out-of-distribution dataset}
Here, we quantify the quality of the uncertainty estimation in unseen environments, where a robust estimate is particularly important.
We utilized the models of EfficientPS (+TS) and our EvPSNet trained on the Cityscapes dataset and deployed them on the Indian Driving dataset.
The results in \tabref{tab:IDD_ECE_met} show that our EvPSNet maintains its performance lead in this case as well.

\section{Conclusions}

In this work, we introduced the novel task of uncertainty-aware panoptic segmentation
and provided two metrics, uPQ and pECE, for evaluating corresponding network performances.
We also introduced several strong baselines by combining state-of-the-art panoptic segmentation networks with sampling-free uncertainty estimation techniques.
Our proposed EvPSNet architecture outperforms all the baselines on the proposed metrics, and on the standard PQ.
Moreover, we provide extensive evaluations and studies to assist future works in adapting their networks to become uncertainty-aware.
We believe this will motivate future works, in the field of reliable and probabilistic holistic scene understanding,
for example by utilizing different versions of evidential deep learning \cite{ulmer2021survey},
which is crucial for the safe operation of autonomous vehicles.







\bibliographystyle{IEEEtran}
\bibliography{main.bib}
\clearpage
\renewcommand{\baselinestretch}{1}
\setlength{\belowcaptionskip}{0pt}

\begin{strip}
\begin{center}
\vspace{-5ex}
\textbf{\LARGE \bf
Uncertainty-aware Panoptic Segmentation
} \\
\vspace{2ex}

\Large{\bf- Supplementary Material -}\\
\vspace{0.4cm}
\normalsize{Kshitij Sirohi, Sajad Marvi, Daniel B\"uscher and Wolfram Burgard}
\end{center}
\end{strip}

\setcounter{section}{0}
\setcounter{equation}{0}
\setcounter{figure}{0}
\setcounter{table}{0}
\setcounter{page}{1}
\makeatletter

\renewcommand{\thesection}{S.\arabic{section}}
\renewcommand{\thesubsection}{S.\arabic{subsection}}
\renewcommand{\thetable}{S.\arabic{table}}
\renewcommand{\thefigure}{S.\arabic{figure}}

%


\normalsize

\begin{figure}
\captionsetup[subfigure]{aboveskip=1ex,belowskip=1ex}
\begin{subfigure}{0.495\linewidth}
 \includegraphics[width=1.0\linewidth]{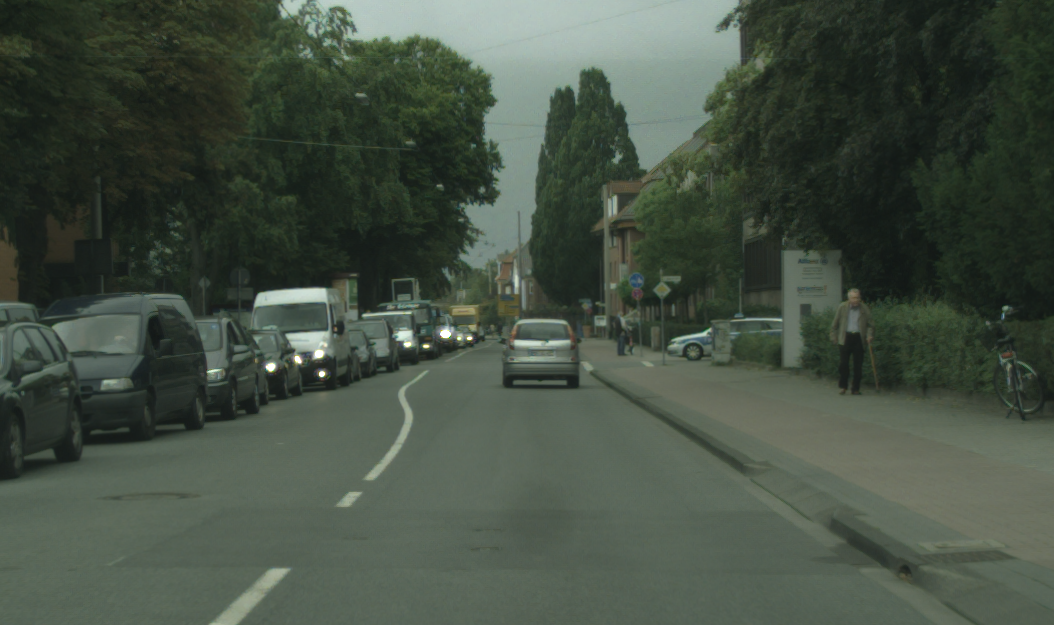}
 \subcaption{Input Image}\label{fig:app_intro_input_image}
\end{subfigure}
\begin{subfigure}{0.495\linewidth}
 \includegraphics[width=1.0\linewidth]{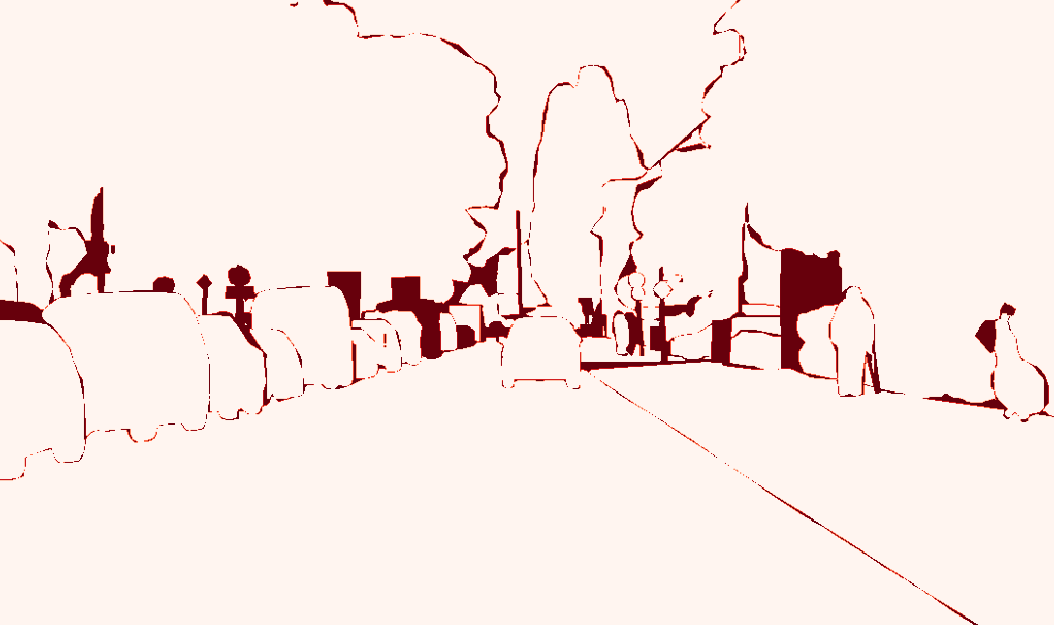}
 \subcaption{Error Map}\label{fig:app_intro_error_map}
\end{subfigure}
\caption{Input image and the corresponding error map for the predictions in \figref{fig:examples} of the main paper.}
\label{fig:appendix_intro}
\end{figure}

In this supplementary material, we provide additional ablation studies on our design choices for the activation function, loss functions, performance on the out-of-distribution (OOD) Indian Driving dataset.
Additional qualitative examples are shown for the Cityscapes data.
Similar to the main paper, all results presented in this appendix are calculated from our testing (= default-validation) sets unless stated otherwise. 
We trained on the whole training set for the following experiments.
In \figref{fig:appendix_intro} we present the input image and the error map that correspond to \figref{fig:examples} in the main paper.


\section{Additional Ablation Studies}

\subsubsection{Evidential loss choice} \label{sec:losses}

The authors of \cite{sensoy2018evidential} suggest three versions of loss functions for evidential deep learning for classification.
The first includes obtaining a negated logarithm of the prior likelihood as defined in \eqref{eq:log_loss_sem}, also known as the Type-II maximum likelihood technique.
The second one computes the Bayes risk for the class predictor, and for cross-entropy loss, the final Bayes risk is similar to \eqref{eq:log_loss_sem} but just replacing log by digamma.
The final version is the mean squared error (MSE), which calculates the sum of squares error between the ground truth one-hot encoding and the predicted probabilities. 
The authors suggest to use the MSE loss and left the study of other losses for future work.

Hence, we replaced our respective segmentation and classification loss functions in the semantic and instance segmentation heads with the MSE loss for comparison.
From \tabref{tab:loss_types} we can see that the pECE value for the MSE loss is significantly worse than ours.
Moreover, the performance of PQ\textsuperscript{Th} is significantly affected.

In our next experiment, we utilize the MSE loss only for the semantic segmentation head, while training the instance segmentation head with the original loss as in \cite{mohan2021efficientps}.
This variant is represented by MSE* and leads to better pECE and PQ\textsuperscript{Th} values.
Thus, we deduce that the MSE loss is not suitable for the instance segmentation head, or at least requires modifications.

Our log loss performs the best out of the three versions in our multi-class segmentation setting.
Please note that we did not include the Lov{\'a}sz evidential loss for semantic segmentation here, allowing for a fair comparison of only the evidential losses.

We conclude that the MSE loss, even though suggested for the classification task and used in semantic segmentation task for a few classes, is unsuitable when the number of classes is large and the data is unbalanced.
While MSE loss aims to regress probabilities, it comes with the disadvantage of not taking care of finding the best class, which other losses aim for.

\begin{table*}
\begin{center}
\caption{Comparison of loss functions on the Cityscapes data. All scores are in [\%].}
\label{tab:loss_types}
\footnotesize
\begin{tabular}
{l|cccc|cccc|cccc|c}
\toprule
Method & uPQ & PQ  & SQ & RQ & uPQ\textsuperscript{Th}& PQ\textsuperscript{Th} & SQ\textsuperscript{Th} & RQ\textsuperscript{Th} & uPQ\textsuperscript{St} & PQ\textsuperscript{St} & SQ\textsuperscript{St} & RQ\textsuperscript{St} & pECE  \\
\midrule
MSE & 33.0 & 55.5 & 72.3 & 67.5 & 29.9&41.5  & 59.1 & 52.5 & 33.1 &65.6  &81.9  & 78.5 &   40.4  \\
MSE* & 37.6 & 56.2 & 78.7 & 69.8 & 35.6 & 52.7  & 77.7 & 67.7 & 40.3 &58.7  &79.5  & 71.4 &   33.1  \\
Digamma &$46.5$& $61.3$&$\mathbf{81.5}$ &$74.2$ &$36.5$ &$52.4$  &$\mathbf{80.4}$ &$64.7$  &$54.5$ &$67.8$  &$\mathbf{82.2}$  &$81.1$ & $24.1$   \\
\midrule
Log & $\mathbf{50.0}$ & $\mathbf{63.4}$ & $81.2$ & $\mathbf{77.1}$ & $\mathbf{42.4}$& $\mathbf{55.8}$ & $79.8$ & $\mathbf{69.6}$ & $\mathbf{55.7}$& $\mathbf{68.9}$ & $\mathbf{82.2}$ & $\mathbf{82.5}$ & $\mathbf{21.2}$ \\
\bottomrule
\end{tabular}
\end{center}
\end{table*}

\subsubsection{Activation function} \label{sec:reluvsoftplus}

The activation function at the end of the network calculates the evidence for a particular class provided by the network.
Other methods commonly utilize ReLU as the activation function.
However, we argue that this leads to the loss of small evidence provided by the network for segmentation tasks, especially at the beginning of training.
Thus we utilize softplus activation in our network, which keeps the smaller values rather than nullifying them.
As listed in \tabref{tab:evidence}, replacing ReLU with softplus leads to an increase of 0.8 percent in PQ, 0.4 percent in PQ\textsuperscript{Th}, and 1.1 percent in PQ\textsuperscript{St}.
Similarly we gain 1.1 percent in uPQ, 0.9 in uPQ\textsuperscript{Th} and 1.1 percent in uPQ\textsuperscript{St}.
The softplus activation also leads to a gain of 0.6 percent in pECE, hence provides more accurate uncertainty estimate.
Please note, that we train the network only with evidential losses without the proposed Lov{\'a}sz evidential loss here to clearly analyze the activation function.

\begin{table*}
\begin{center}
\caption{Comparison of evidence functions on the Cityscapes data. All scores are in [\%].}
\label{tab:evidence}
\footnotesize
\begin{tabular}
{l|cccc|cccc|cccc|c}
\toprule
Method & uPQ & PQ  & SQ & RQ & uPQ\textsuperscript{Th}& PQ\textsuperscript{Th} & SQ\textsuperscript{Th} & RQ\textsuperscript{Th} & uPQ\textsuperscript{St} & PQ\textsuperscript{St} & SQ\textsuperscript{St} & RQ\textsuperscript{St} & pECE  \\
\midrule
ReLU & $48.9$  &$62.6$ &  $81.3$ & $76.0$ & $41.8$  &$55.4$ & $\mathbf{80.0}$ & $68.9$& $54.3$ & $67.9$ & $\mathbf{82.2}$ & $81.2$ & $21.9$  \\
\midrule
softplus & $\mathbf{50.0}$ & $\mathbf{63.4}$ & $81.2$ & $\mathbf{77.1}$ & $\mathbf{42.4}$& $\mathbf{55.8}$ & $79.8$ & $\mathbf{69.6}$ & $\mathbf{55.7}$& $\mathbf{68.9}$ & $\mathbf{82.2}$ & $\mathbf{82.5}$ & $\mathbf{21.2}$ \\
\bottomrule
\end{tabular}
\end{center}
\end{table*}

\subsubsection{Lov{\'a}sz evidential loss} \label{sec:Lovaszloss}

While the evidential log loss aims at better uncertainty estimation, the Lov{\'a}sz softmax loss aims at improving the segmentation quality by utilizing the softmax-based probabilities.
We combine the better uncertainty and probability estimate from evidential log loss with the Lov{\'a}sz loss to optimize both uncertainty estimation and the IoU maximization.
In \tabref{tab:evidence_loss} we demonstrate the advantage of utilizing our proposed Lov{\'a}sz evidential loss function.
Our loss function gains 1.7\% on uPQ and 0.7\% on the PQ.
It is interesting to note that even though the loss is only utilized in the semantic segmentation head, it still improves both uPQ\textsuperscript{Th} and PQ\textsuperscript{Th}.
Moreover, better segmentation quality contributes to better boundary estimation, increasing the pECE metric performance.

\begin{table*}
\begin{center}
\caption{Comparison of different proposed loss function on the Cityscapes data. All scores are in [\%].}
\label{tab:evidence_loss}
\footnotesize
\begin{tabular}
{l|cccc|cccc|cccc|c}
\toprule
Method & uPQ & PQ  & SQ & RQ & uPQ\textsuperscript{Th}& PQ\textsuperscript{Th} & SQ\textsuperscript{Th} & RQ\textsuperscript{Th} & uPQ\textsuperscript{St} & PQ\textsuperscript{St} & SQ\textsuperscript{St} & RQ\textsuperscript{St} & pECE  \\
\midrule
Log & ${50.0}$ & ${63.4}$ & $81.2$ & ${77.1}$ & ${42.4}$& ${55.8}$ & $79.8$ & ${69.6}$ & ${55.7}$& ${68.9}$ & ${82.2}$ & ${82.5}$ & ${21.2}$ \\
\midrule
Lov{\'a}sz evidential (ours) & $\mathbf{51.7}$ & $\mathbf{64.1}$ & $\mathbf{81.4}$ & $\mathbf{77.8}$ & $\mathbf{45.9}$& $\mathbf{56.8}$ & $\mathbf{80.2}$ & $\mathbf{70.6}$ & $\mathbf{56.0}$& $\mathbf{69.5}$ & $\mathbf{82.3}$ & $\mathbf{83.1}$ & $\mathbf{19.4}$ \\
\bottomrule
\end{tabular}
\end{center}
\end{table*}

\section{Additional Qualitative Results}

In this section, we present additional qualitative results on the Cityscapes dataset. Moreover, we also present the out-of-distribution results on the Indian Driving dataset. 

\subsubsection{Cityscapes dataset}

\begin{figure*}
\centering
\footnotesize
\setlength{\tabcolsep}{0.05cm}
{
\renewcommand{\arraystretch}{0.2}
\newcolumntype{M}[1]{>{\centering\arraybackslash}m{#1}}
\begin{tabular}{cM{0.22\linewidth}M{0.22\linewidth}M{0.22\linewidth}M{0.22\linewidth}}
& Input Image & Panoptic Segmentation & Uncertainty Map & Error Map \\
\\
\\
\\
\rotatebox[origin=c]{90}{(a)}& {\includegraphics[width=\linewidth, frame]{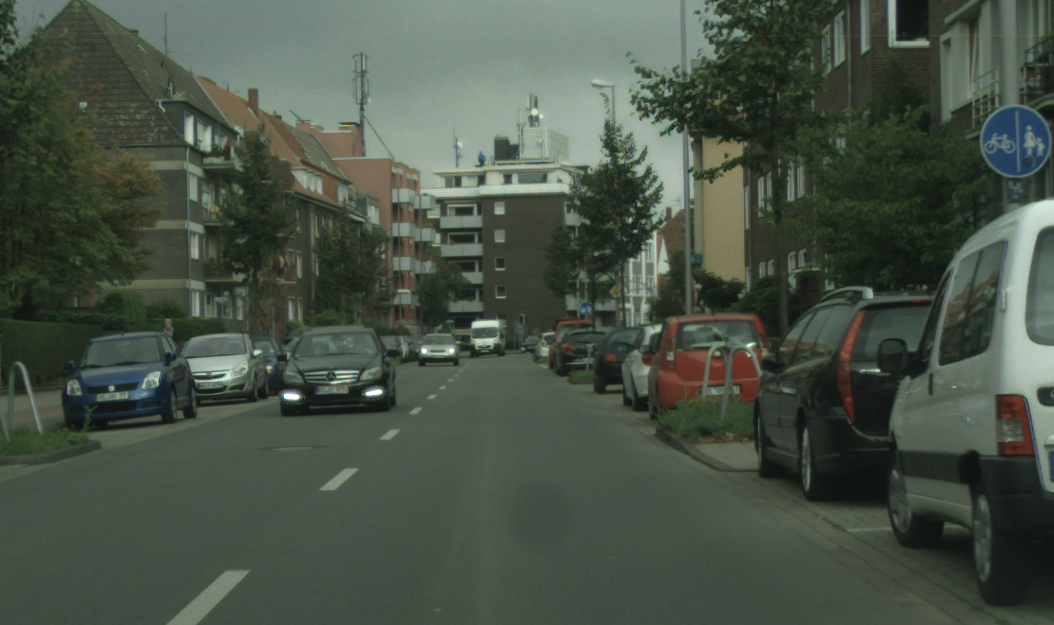}} & {\includegraphics[width=\linewidth, frame]{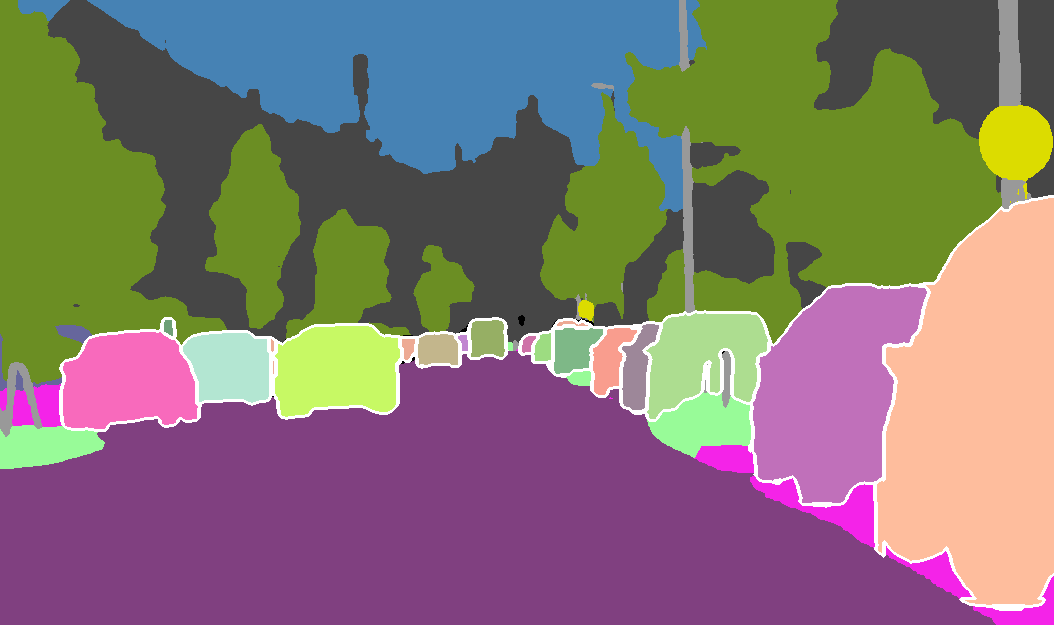}} & {\includegraphics[width=\linewidth, frame]{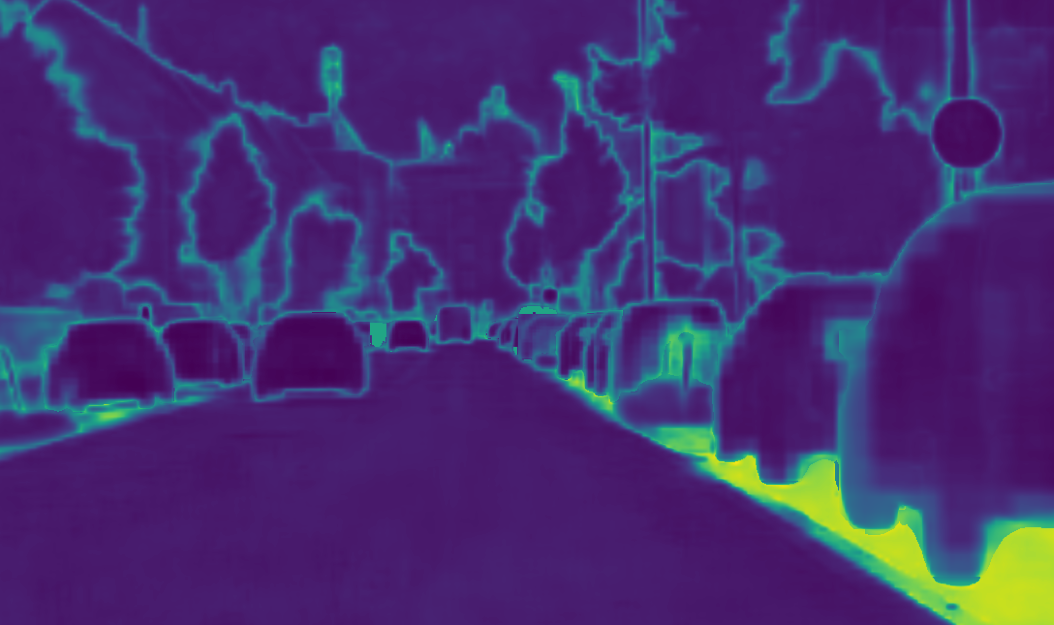}} & {\includegraphics[width=\linewidth, frame]{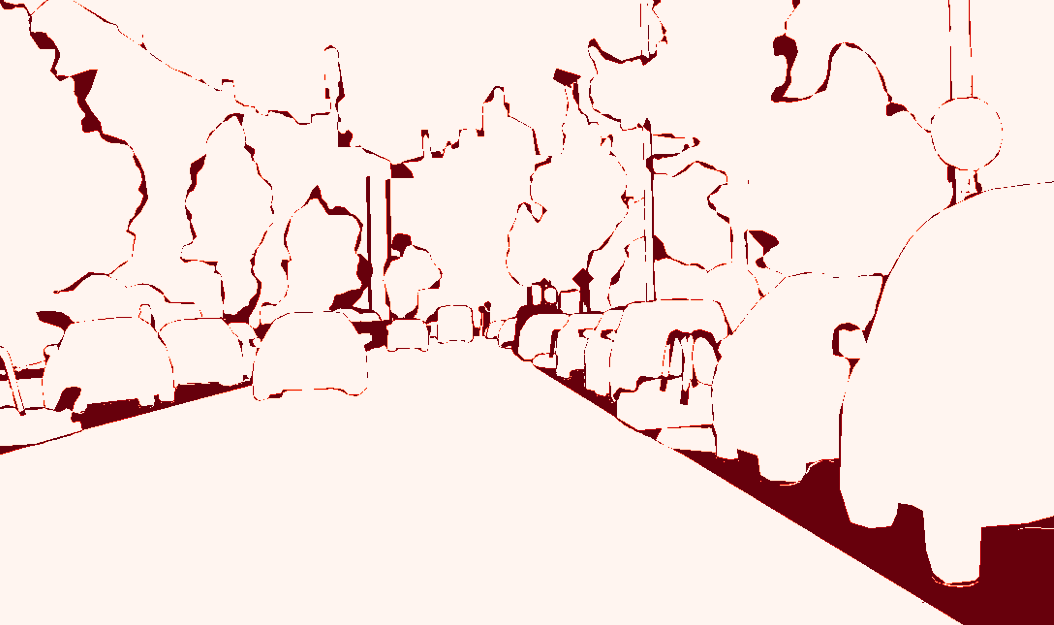}} \\
\\
\rotatebox[origin=c]{90}{(b)}& {\includegraphics[width=\linewidth, frame]{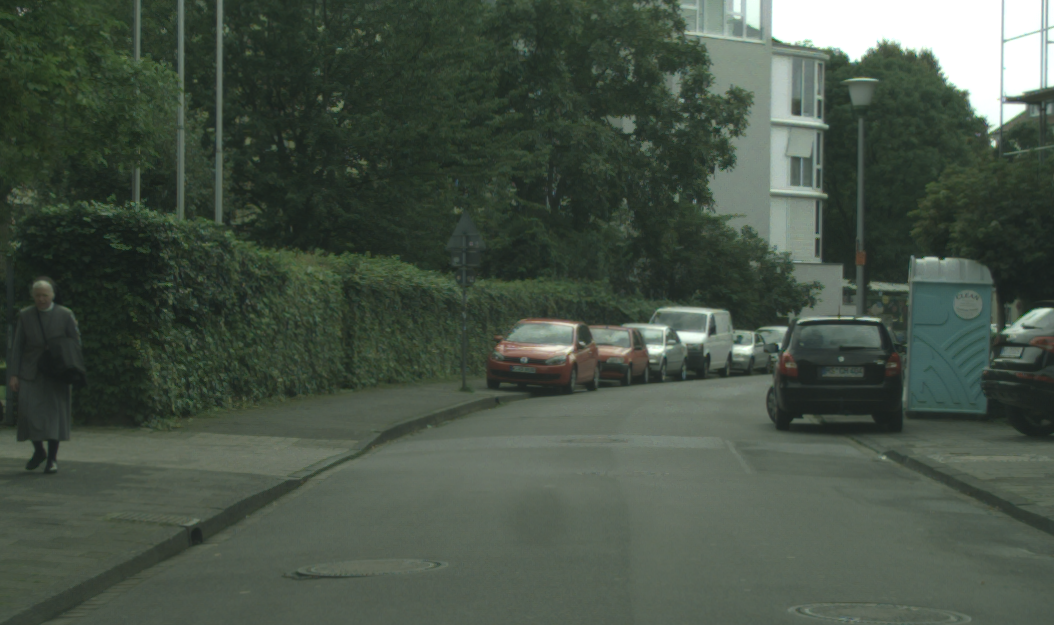}} & {\includegraphics[width=\linewidth, frame]{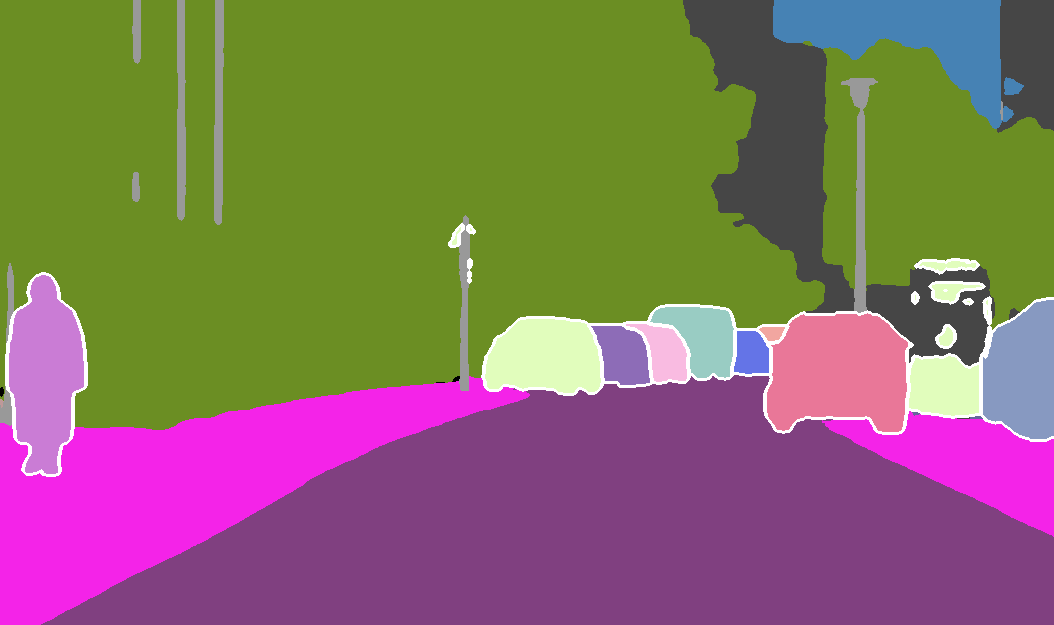}} & {\includegraphics[width=\linewidth, frame]{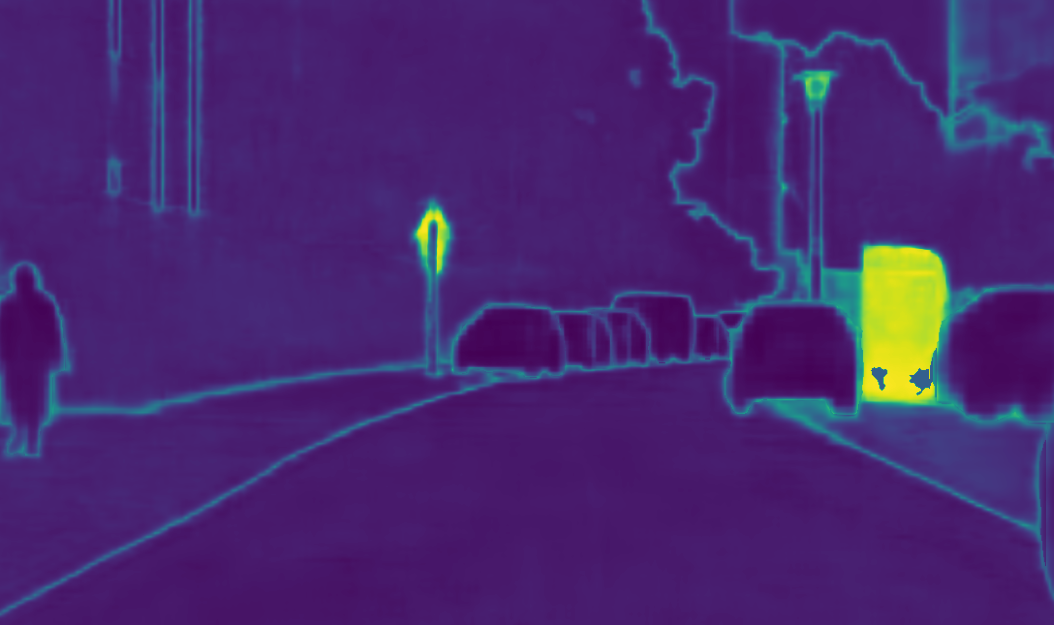}} & {\includegraphics[width=\linewidth, frame]{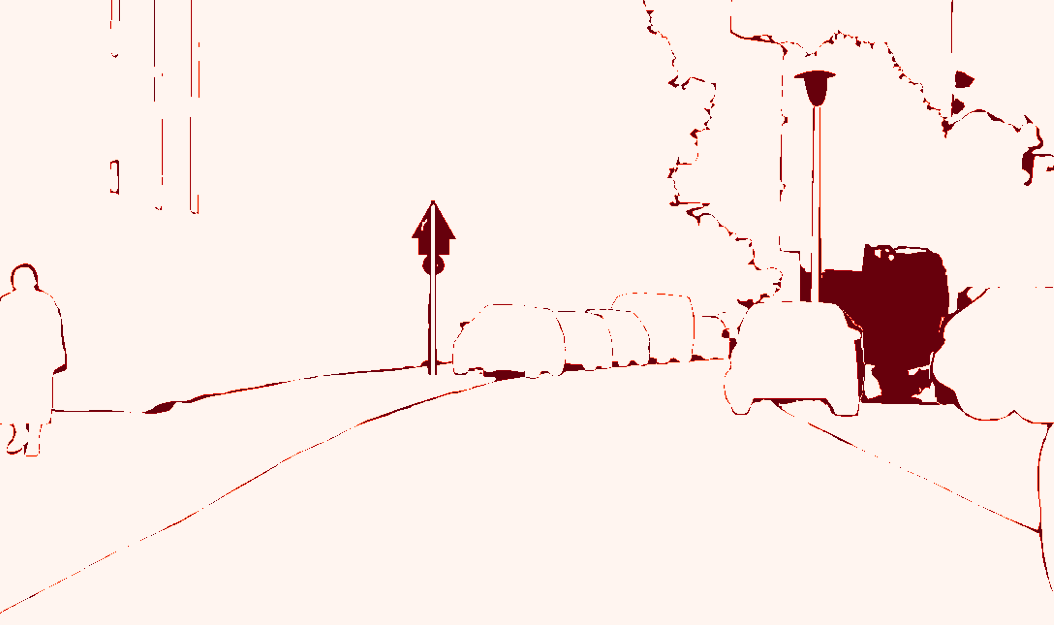}} \\
\\
\rotatebox[origin=c]{90}{(c)}& {\includegraphics[width=\linewidth, frame]{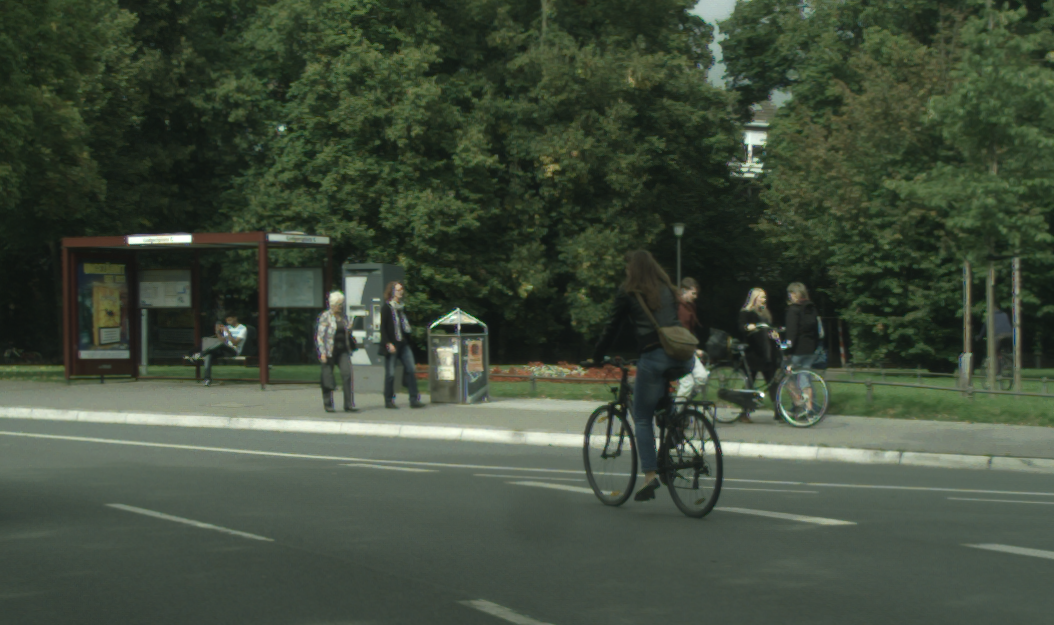}} & {\includegraphics[width=\linewidth, frame]{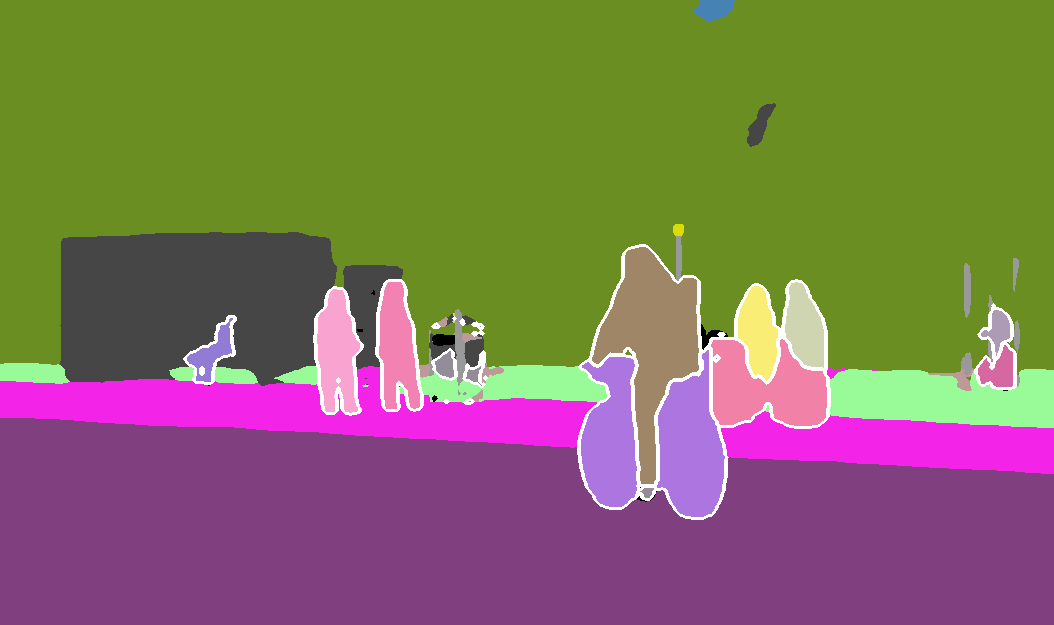}} & {\includegraphics[width=\linewidth, frame]{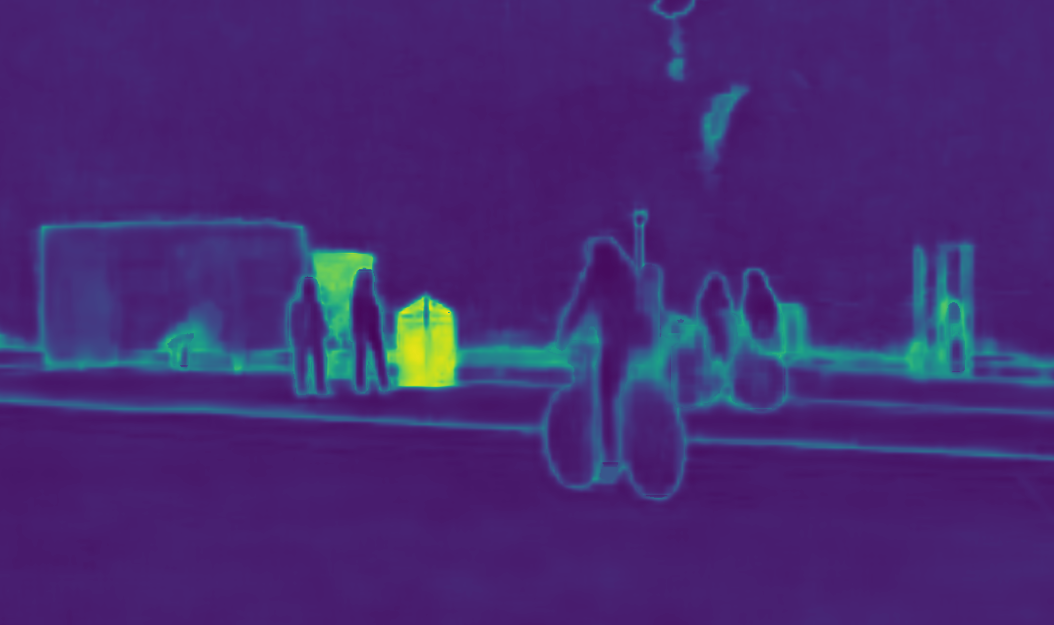}} & {\includegraphics[width=\linewidth, frame]{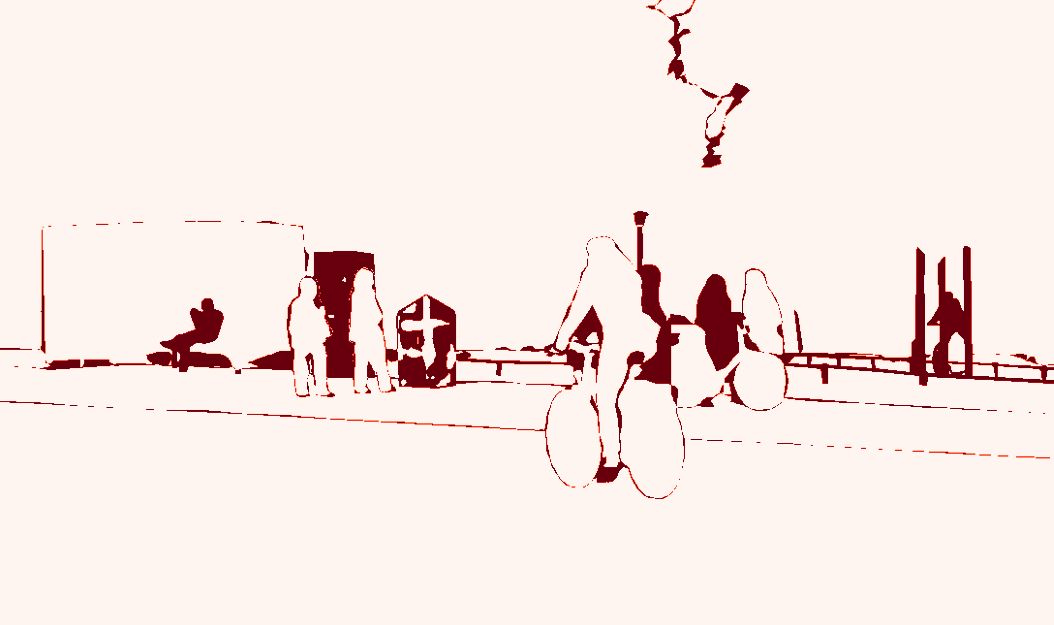}} \\
\\
\rotatebox[origin=c]{90}{(d)}& {\includegraphics[width=\linewidth, frame]{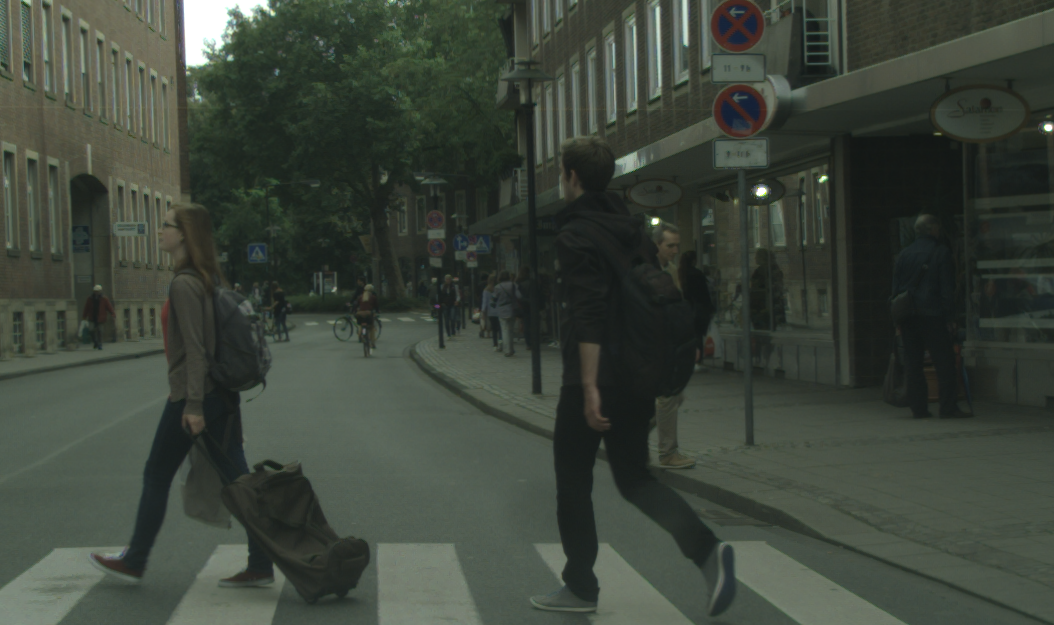}} & {\includegraphics[width=\linewidth, frame]{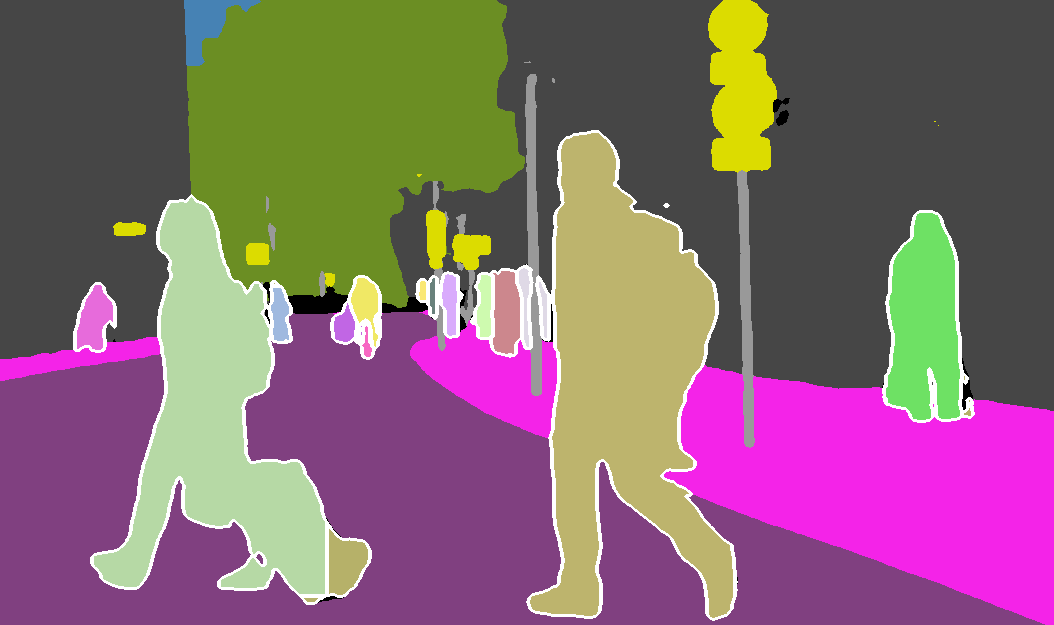}} & {\includegraphics[width=\linewidth, frame]{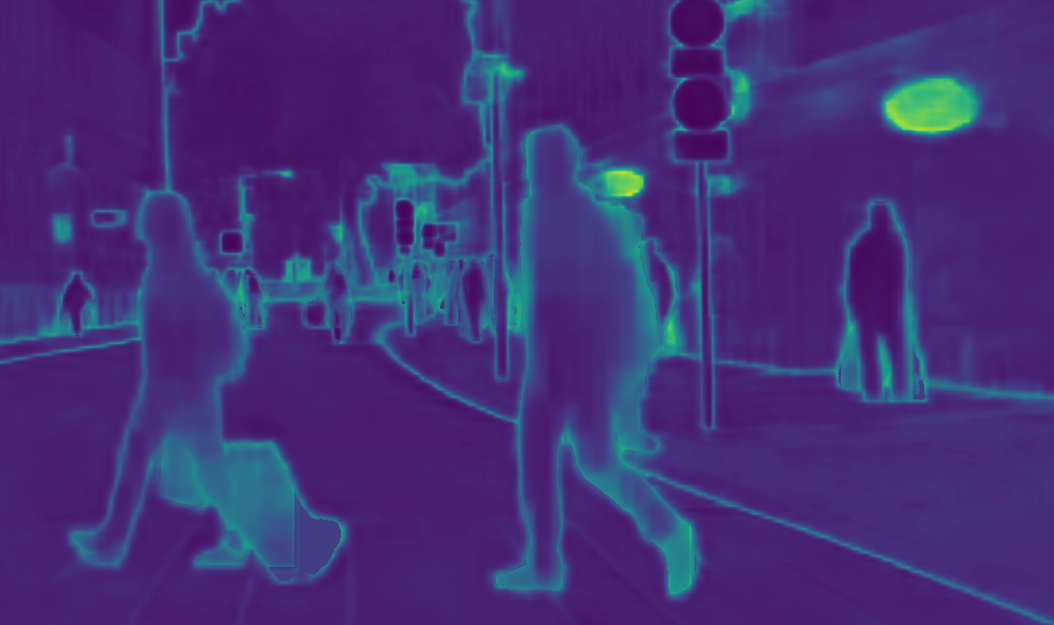}} & {\includegraphics[width=\linewidth, frame]{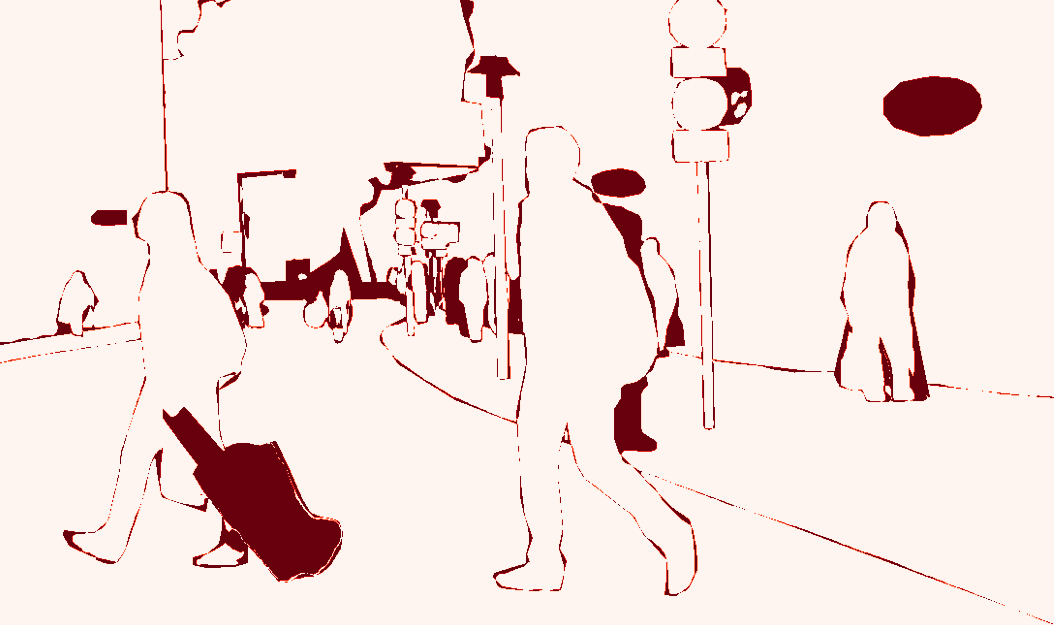}} \\
\end{tabular}
}
\caption{Qualitative results of uncertainty-aware panoptic segmentation by EvPSNet on the Cityscapes data. The visualizations showcase the panoptic segmentation results, corresponding uncertainty, and error maps. The brighter regions in the uncertainty map depict high uncertainty, and the red regions in the error map depict the misclassified pixels by the network. The results showcase that the uncertainty predicted by our network correlates with the errors made by the network.}
\label{fig:additional_city_val}
\vspace{-0.4cm}
\end{figure*}

In \figref{fig:additional_city_val}, we provide additional results on the Cityscapes data. The network is more uncertain about the object it has not been trained for, such as a suitcase and special traffic signs. The network also shows high uncertainty when the surface texture changes, as depicted by high uncertainty and error in \figref{fig:additional_city_val}(a).

\begin{figure*}
\centering
\footnotesize
\setlength{\tabcolsep}{0.05cm}
{
\renewcommand{\arraystretch}{0.2}
\newcolumntype{M}[1]{>{\centering\arraybackslash}m{#1}}
\begin{tabular}{cM{0.22\linewidth}M{0.22\linewidth}M{0.22\linewidth}}
& Input Image & Panoptic Segmentation & Uncertainty Map  \\
\\
\\
\\
\rotatebox[origin=c]{90}{(a)}& {\includegraphics[width=\linewidth, frame]{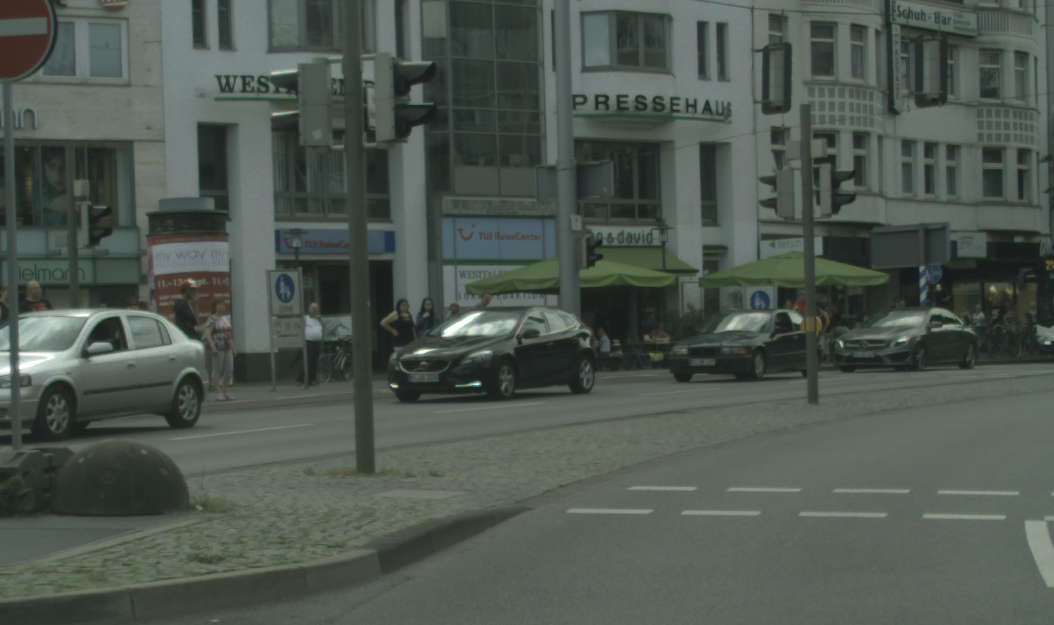}} & {\includegraphics[width=\linewidth, frame]{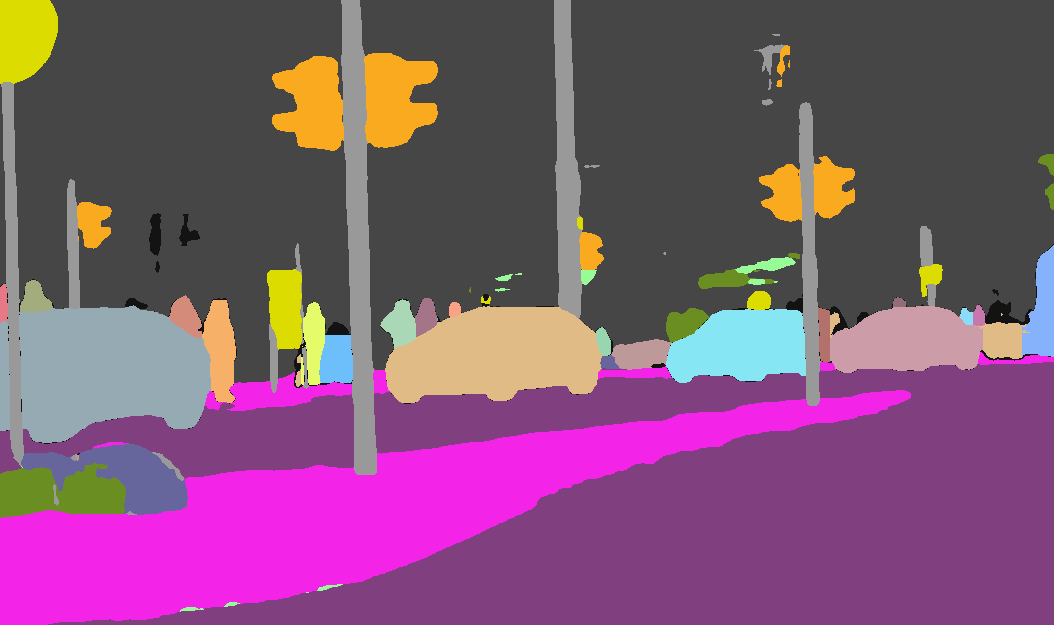}} & {\includegraphics[width=\linewidth, frame]{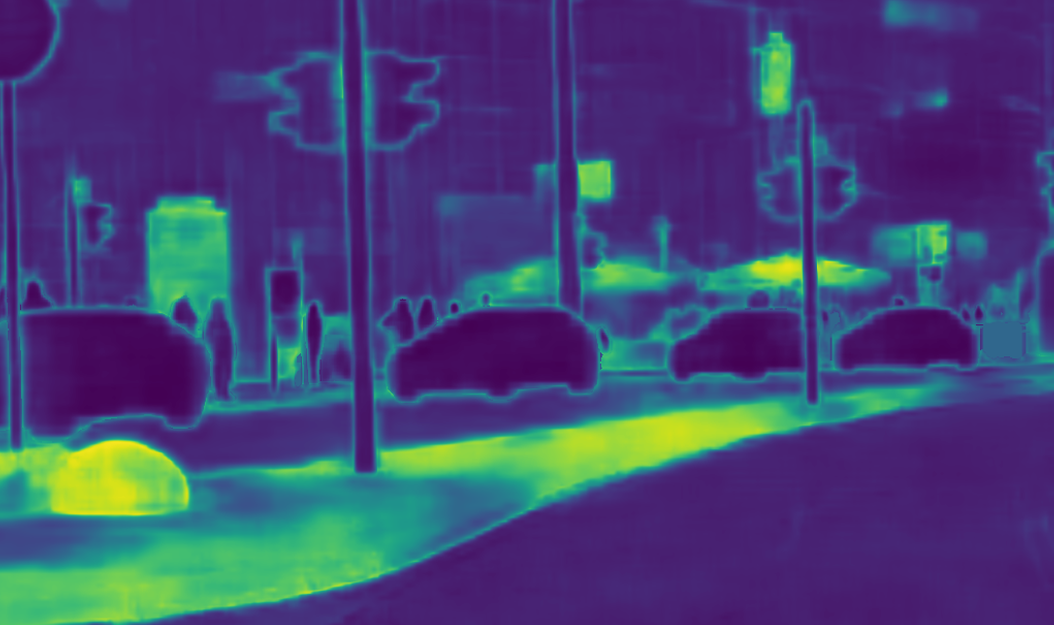}}  \\
\\
\rotatebox[origin=c]{90}{(b)}& {\includegraphics[width=\linewidth, frame]{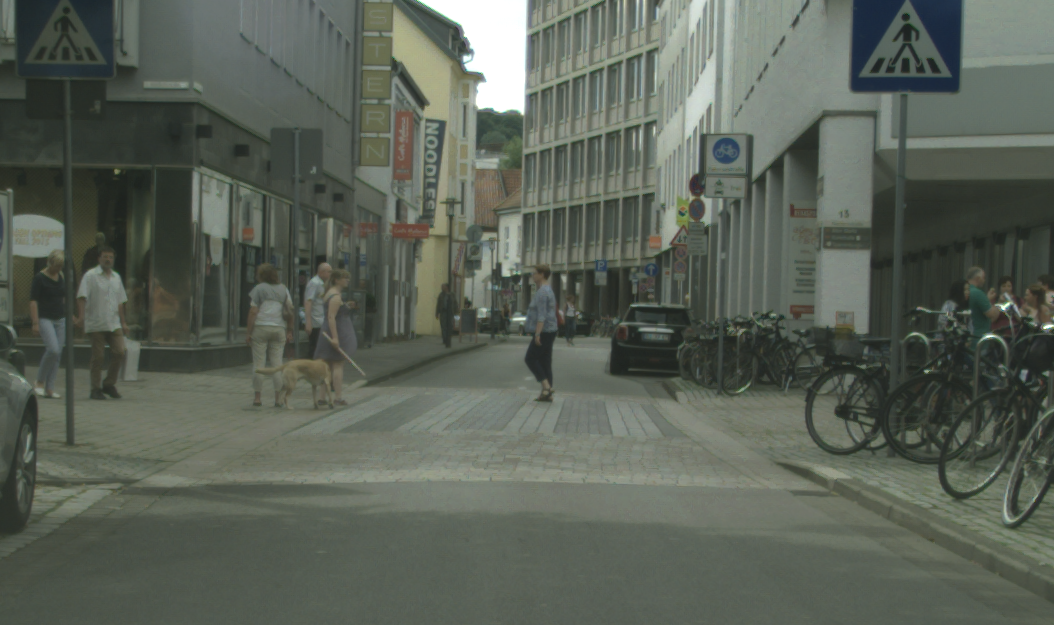}} & {\includegraphics[width=\linewidth, frame]{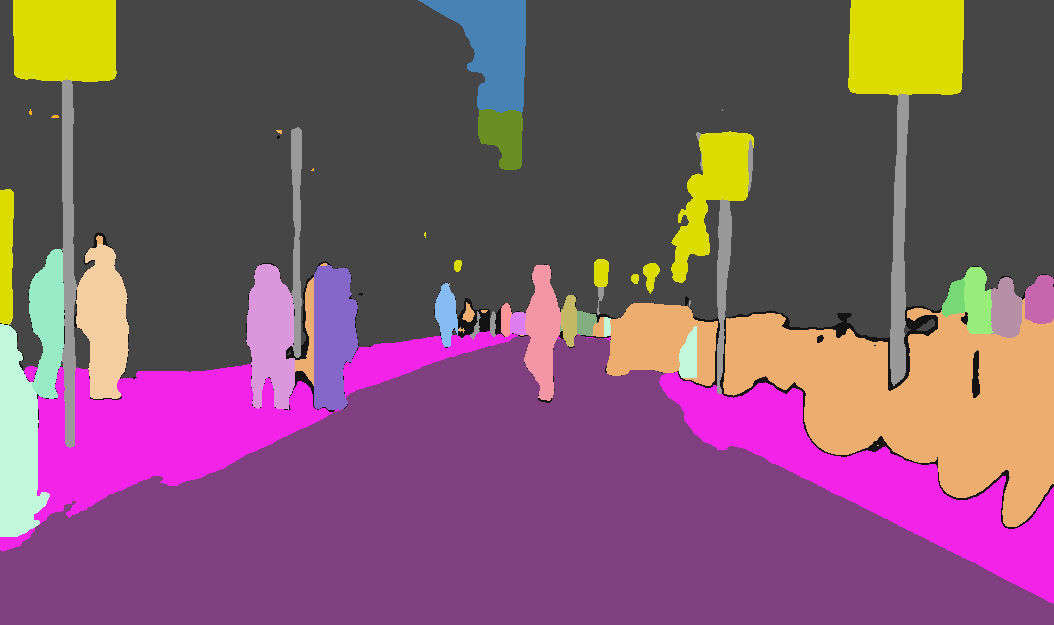}} & {\includegraphics[width=\linewidth, frame]{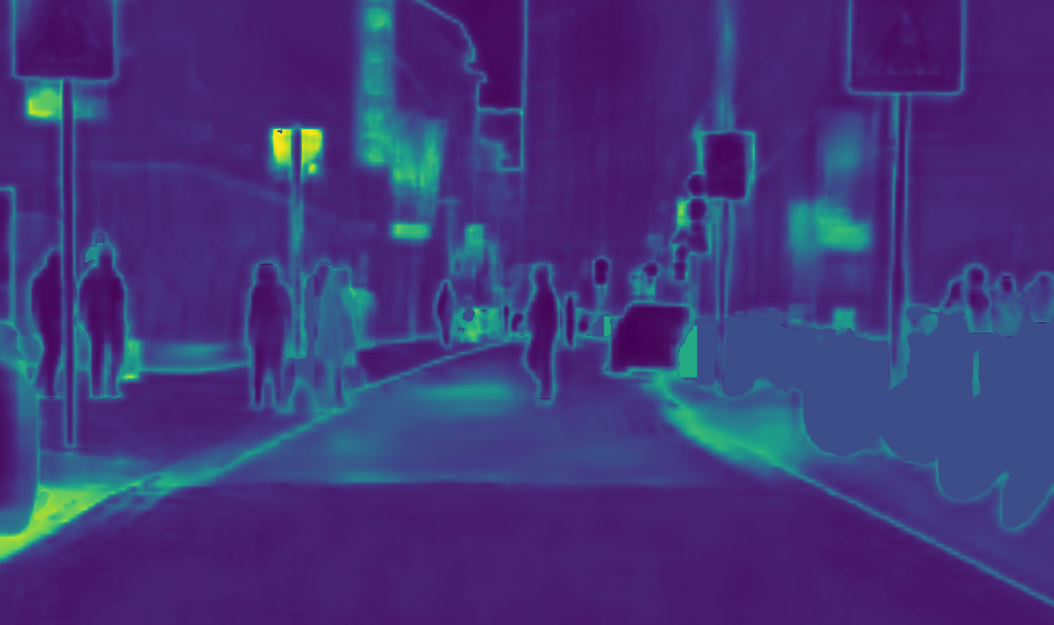}}  \\
\\
\rotatebox[origin=c]{90}{(c)}& {\includegraphics[width=\linewidth, frame]{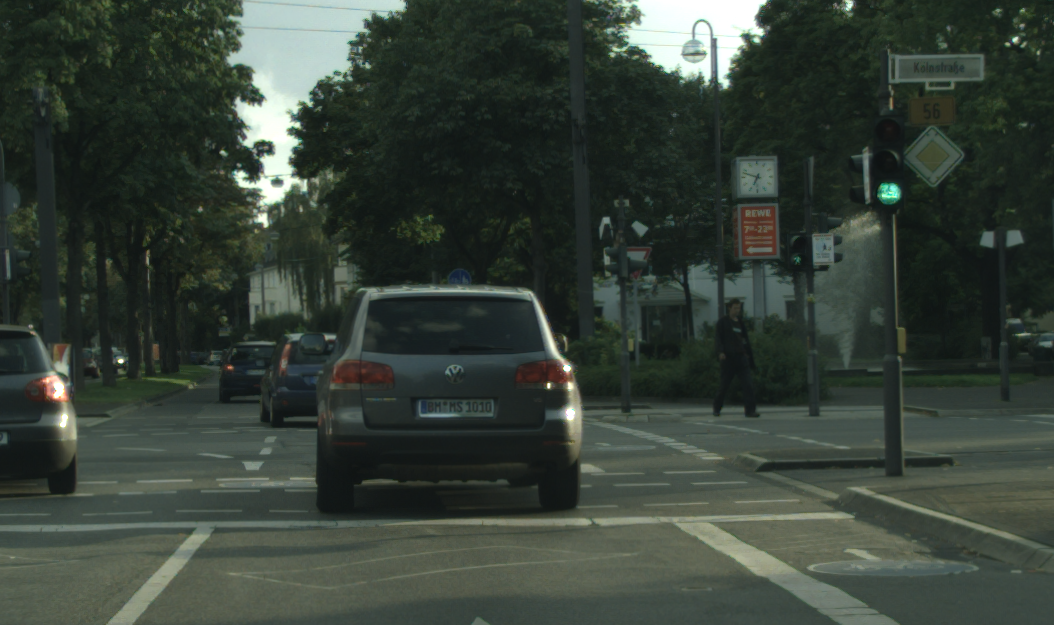}} & {\includegraphics[width=\linewidth, frame]{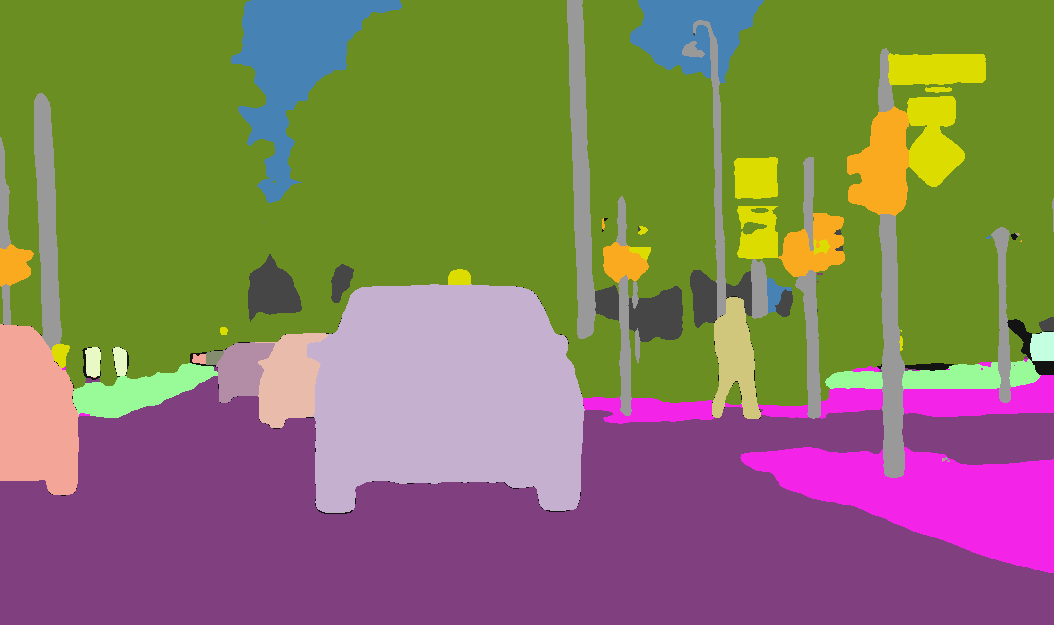}} & {\includegraphics[width=\linewidth, frame]{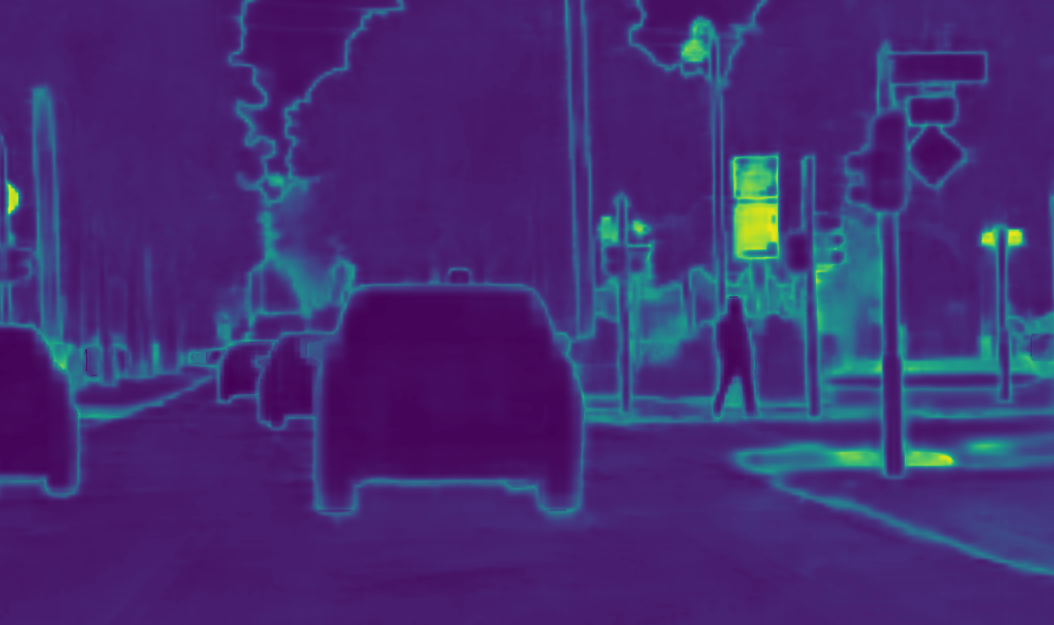}}  \\
\\
\end{tabular}
}
\caption{Qualitative results of uncertainty-aware panoptic segmentation by EvPSNet on the Cityscapes test set. The visualizations showcase the panoptic segmentation results, corresponding uncertainty. The brighter regions in the uncertainty map depict high uncertainty.}
\label{fig:city_test}
\vspace{-0.4cm}
\end{figure*}

We present the results on the original test set of the Cityscapes dataset in \figref{fig:city_test}. We cannot compute the error map because the labels are not provided for the test set. Nonetheless, out network shows predicts high uncertainty for surfaces, such as a street divider in \figref{fig:city_test}(a) and pedestrian crossings in \figref{fig:city_test}(b). It is interesting to note the distinction in the uncertainty map as the uncertainty has the potential to differentiate between different surfaces. In addition, the network predicts the clock and a text advertisement as a traffic sign but also assigns high uncertainty to those in \figref{fig:city_test}(c).

\subsubsection{Indian Driving dataset}

\begin{figure*}
\centering
\footnotesize
\setlength{\tabcolsep}{0.05cm}
{
\renewcommand{\arraystretch}{0.2}
\newcolumntype{M}[1]{>{\centering\arraybackslash}m{#1}}
\begin{tabular}{cM{0.22\linewidth}M{0.22\linewidth}M{0.22\linewidth}M{0.22\linewidth}}
& Input Image & Panoptic Segmentation & Uncertainty Map & Error Map \\
\\
\\
\\
\rotatebox[origin=c]{90}{(a)}& {\includegraphics[width=\linewidth, frame]{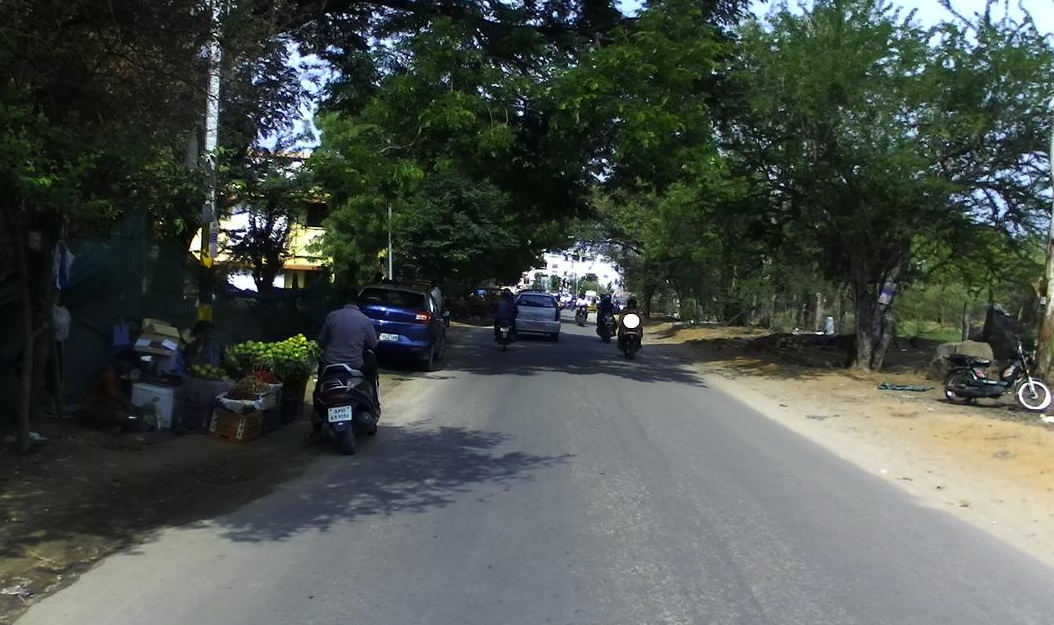}} & {\includegraphics[width=\linewidth, frame]{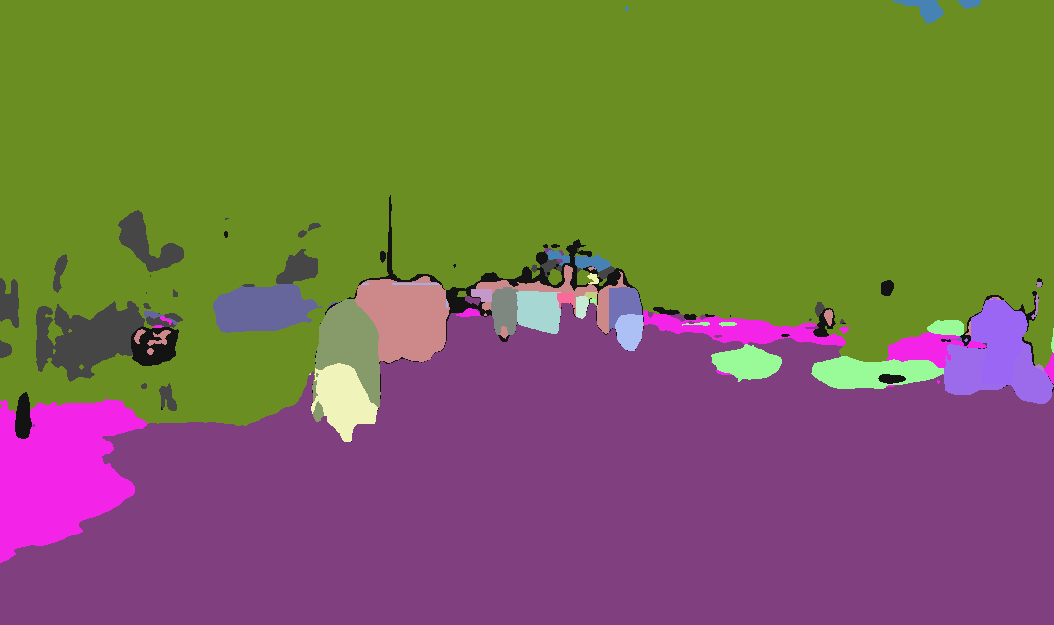}} & {\includegraphics[width=\linewidth, frame]{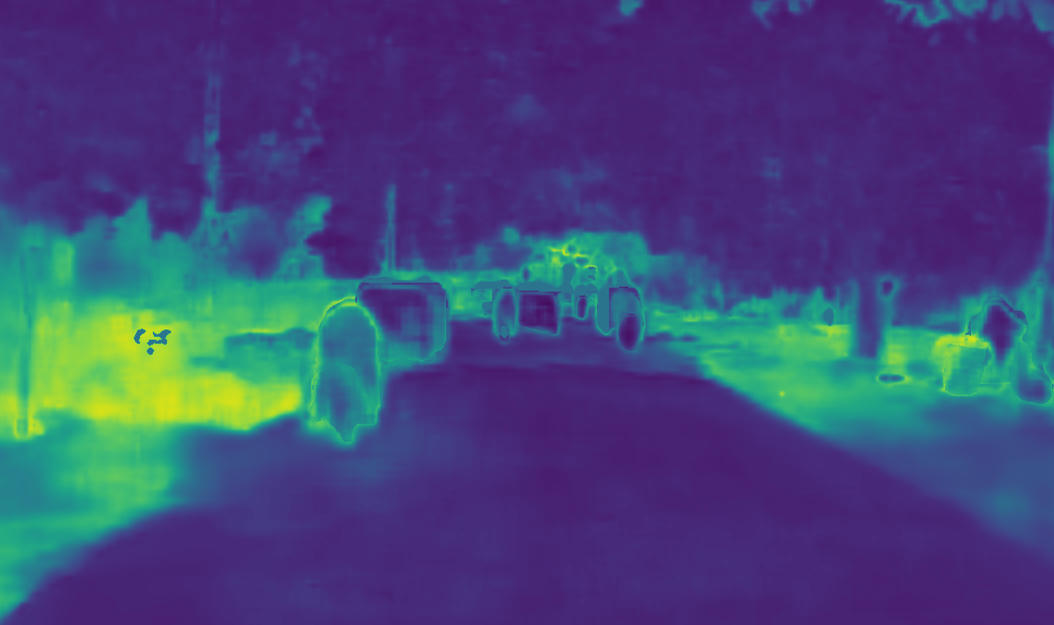}} & {\includegraphics[width=\linewidth, frame]{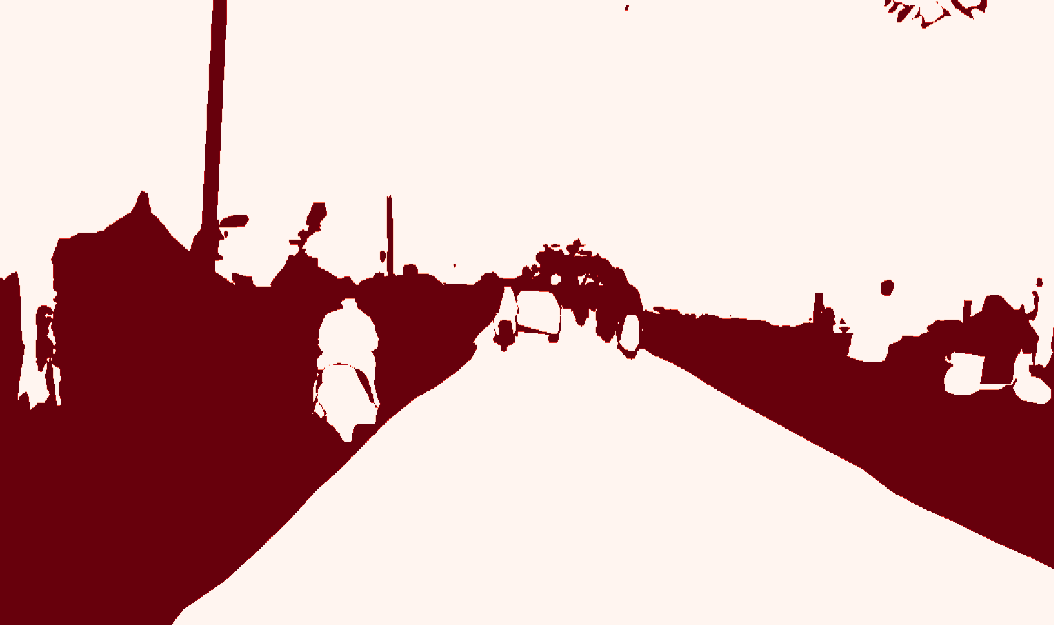}}\\
\\
\rotatebox[origin=c]{90}{(b)}& {\includegraphics[width=\linewidth, frame]{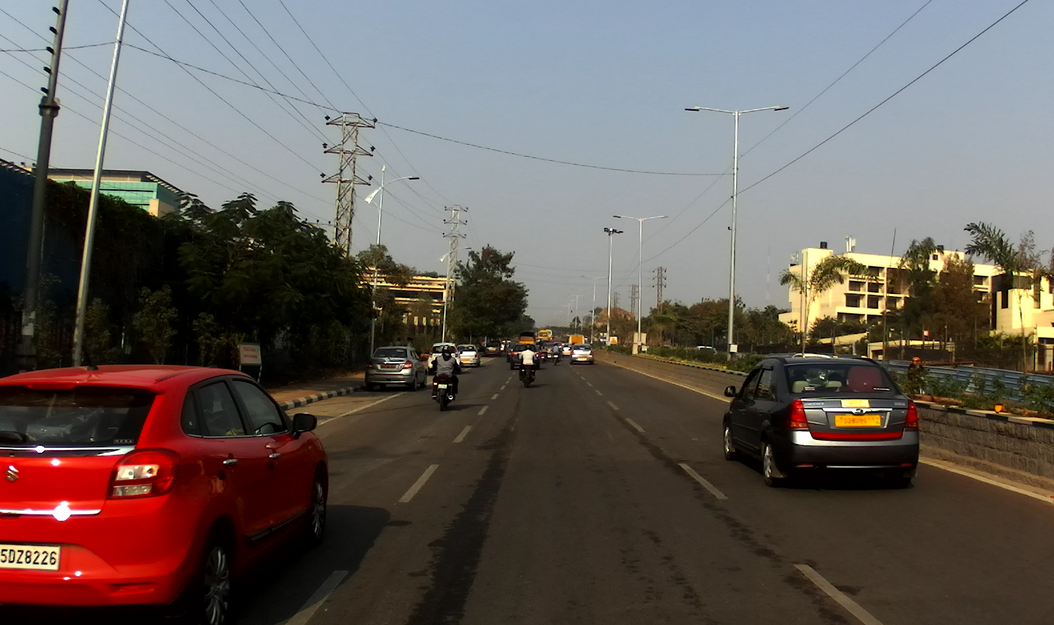}} & {\includegraphics[width=\linewidth, frame]{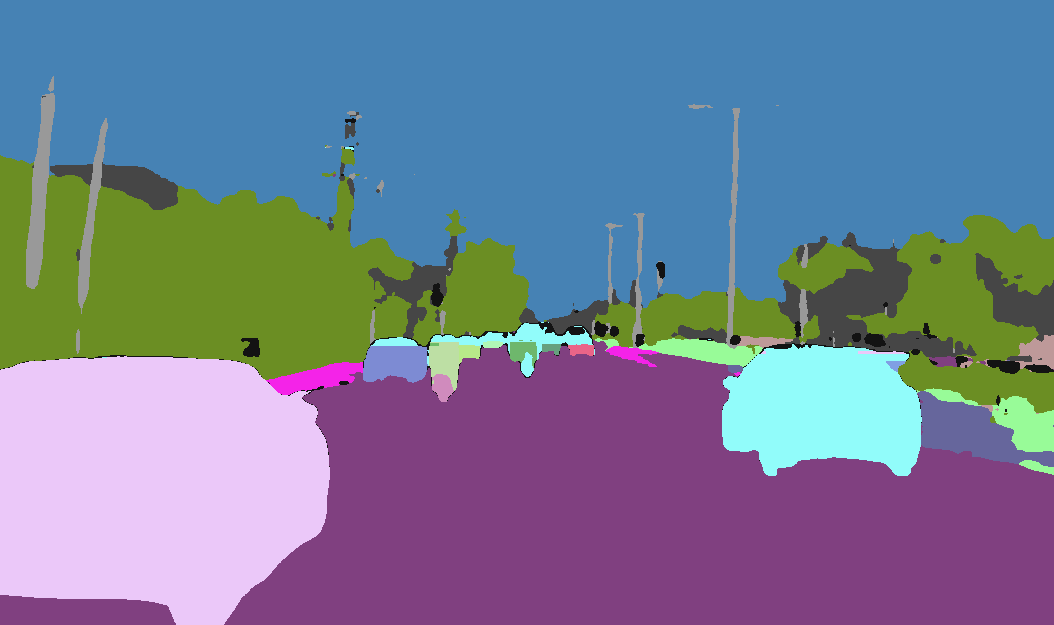}} & {\includegraphics[width=\linewidth, frame]{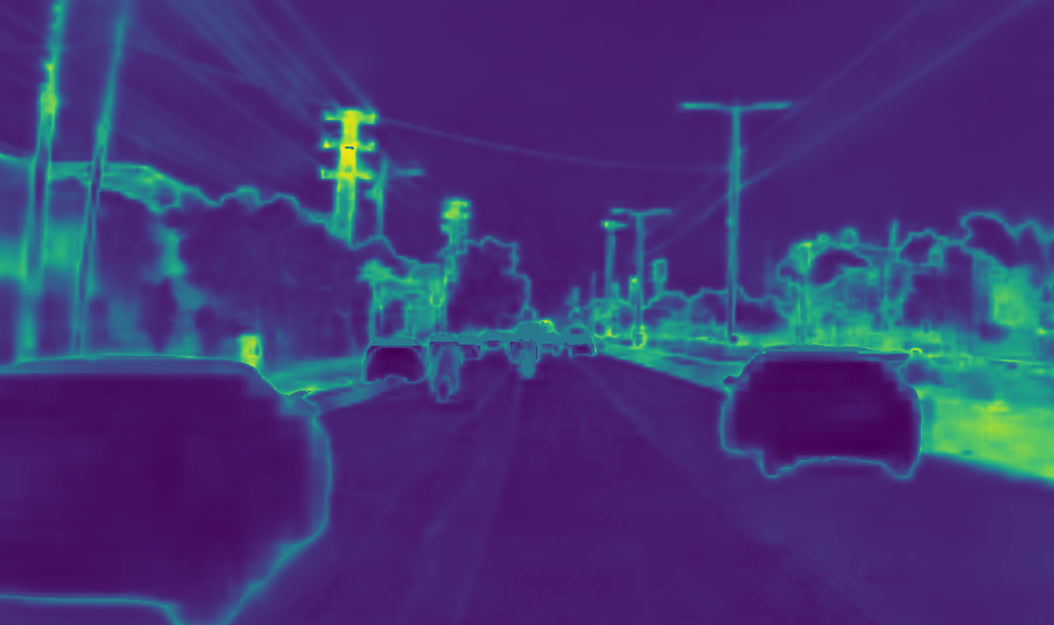}} & {\includegraphics[width=\linewidth, frame]{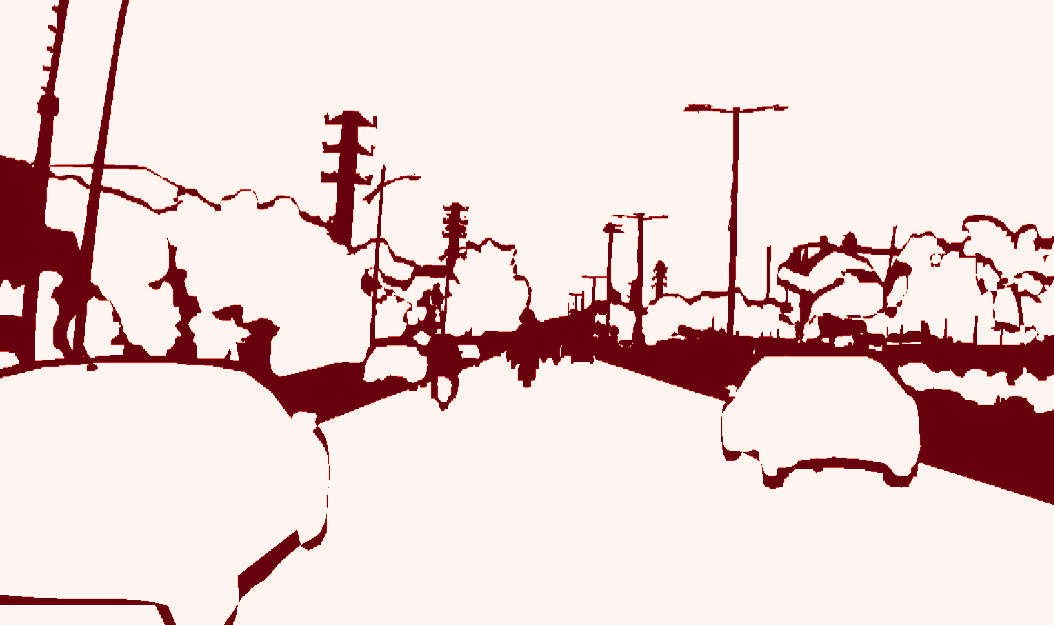}} \\
\\
\rotatebox[origin=c]{90}{(c)}& {\includegraphics[width=\linewidth, frame]{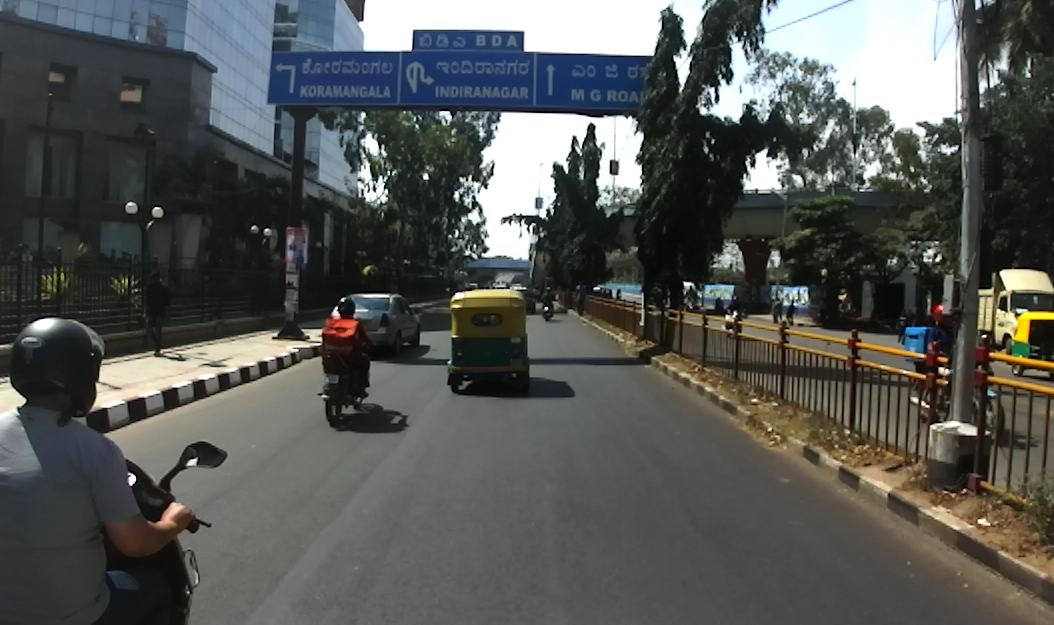}} & {\includegraphics[width=\linewidth, frame]{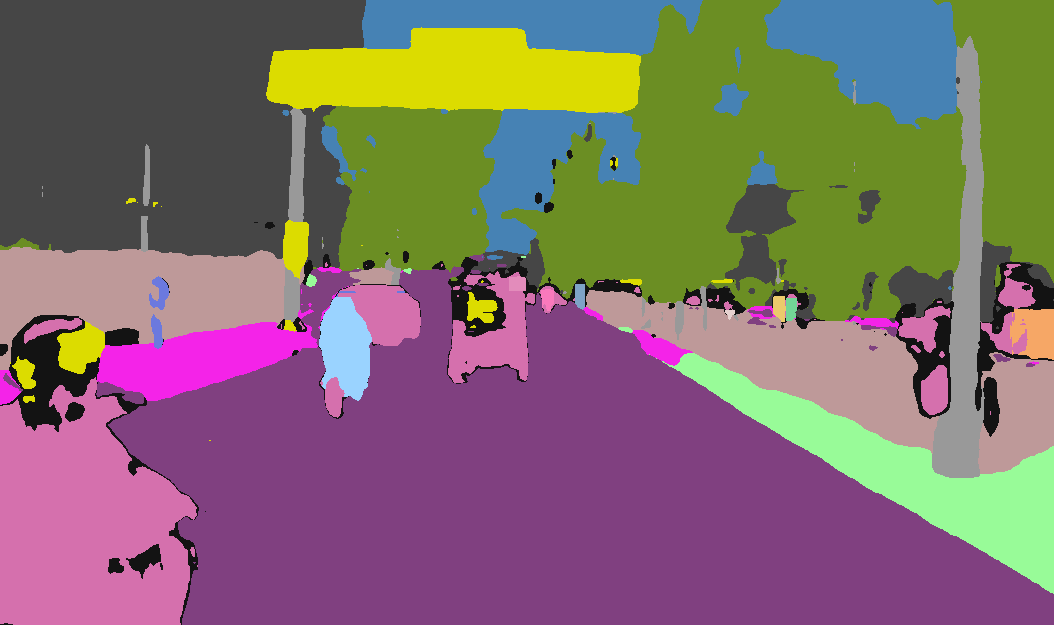}} & {\includegraphics[width=\linewidth, frame]{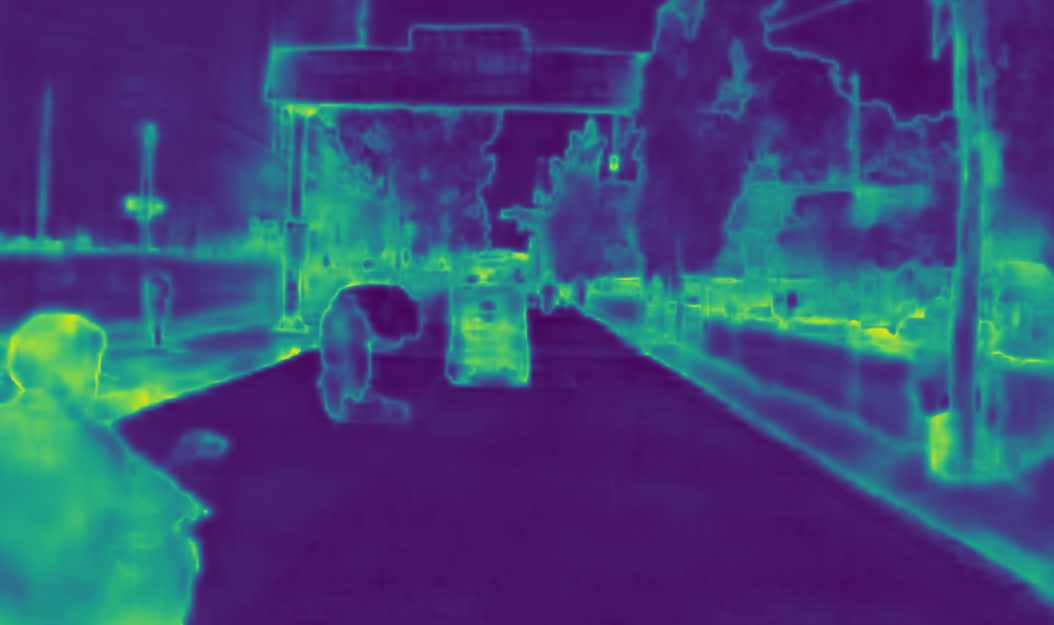}} & {\includegraphics[width=\linewidth, frame]{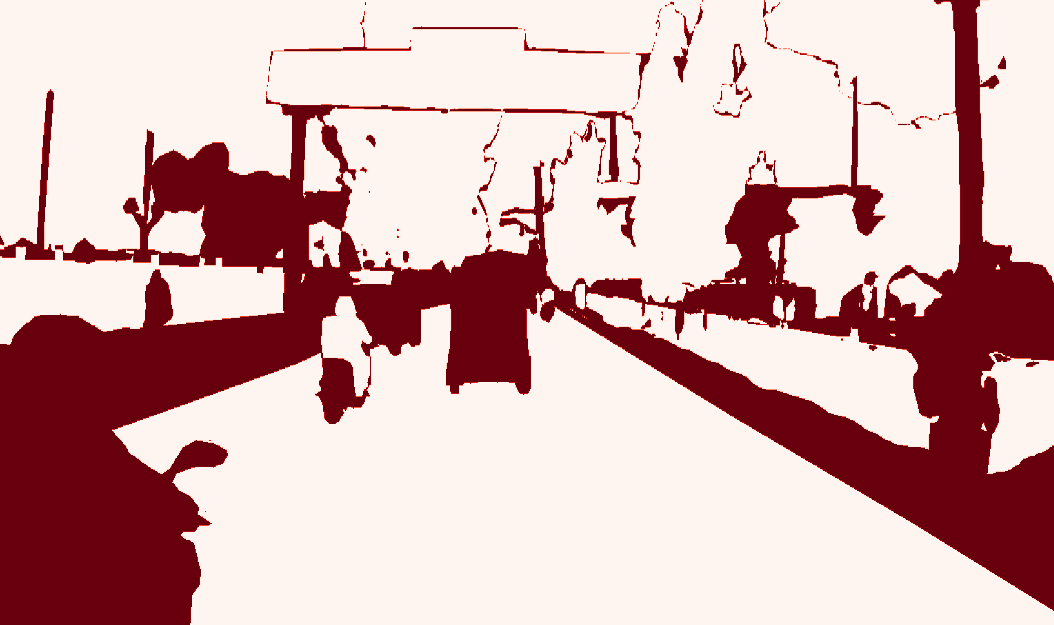}} \\
\\
\rotatebox[origin=c]{90}{(d)}& {\includegraphics[width=\linewidth, frame]{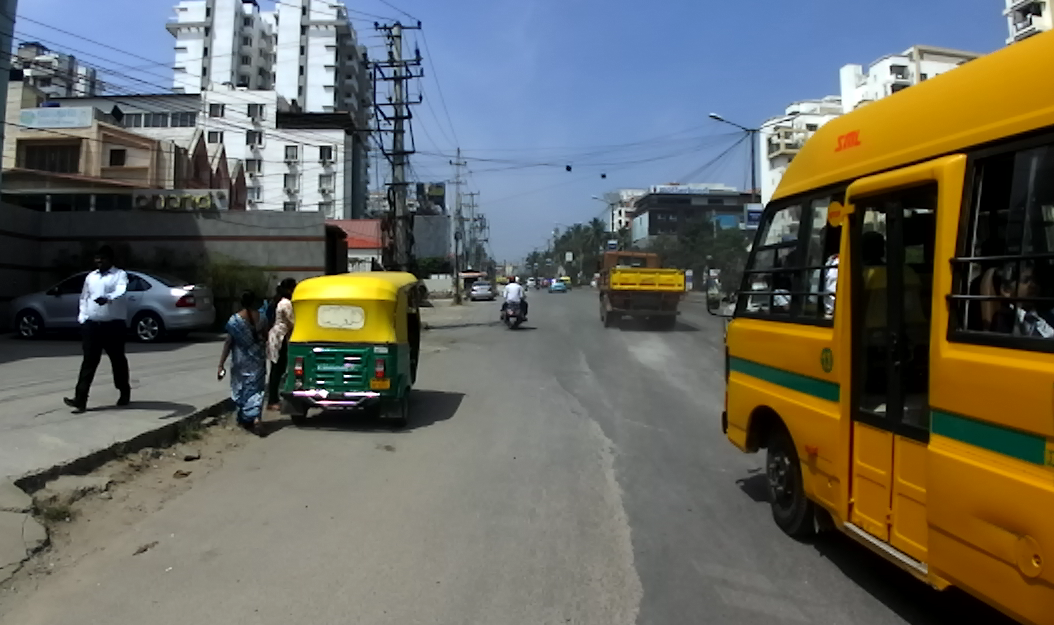}} & {\includegraphics[width=\linewidth, frame]{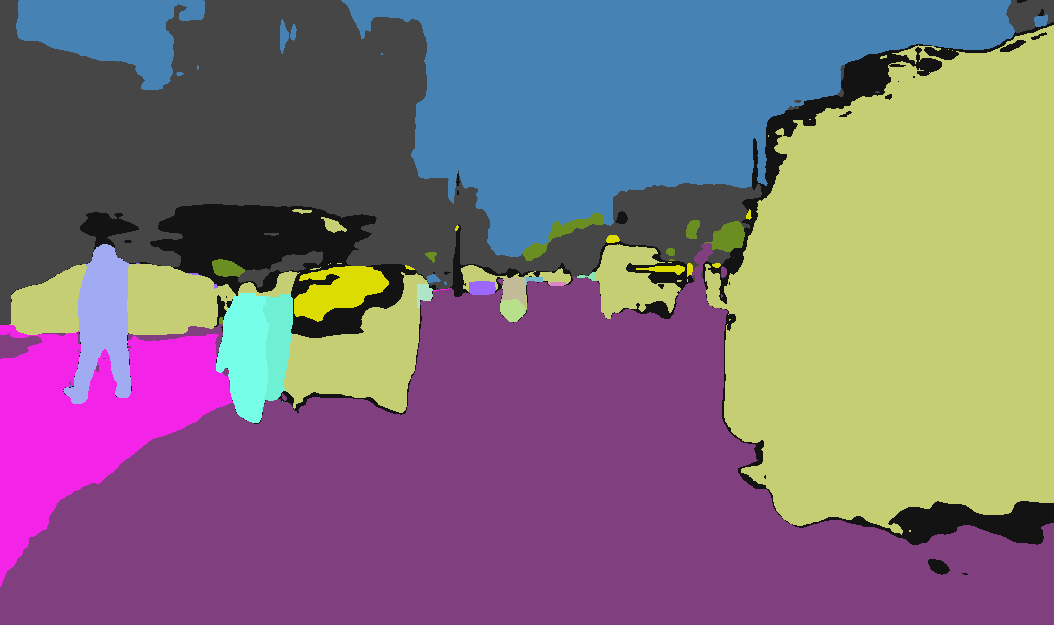}} & {\includegraphics[width=\linewidth, frame]{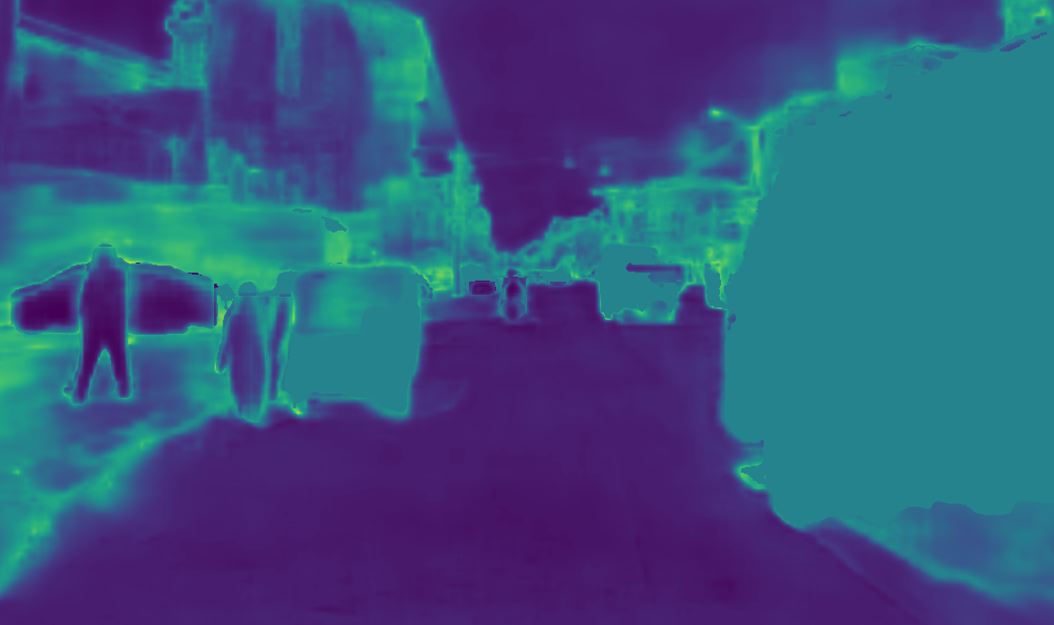}} & {\includegraphics[width=\linewidth, frame]{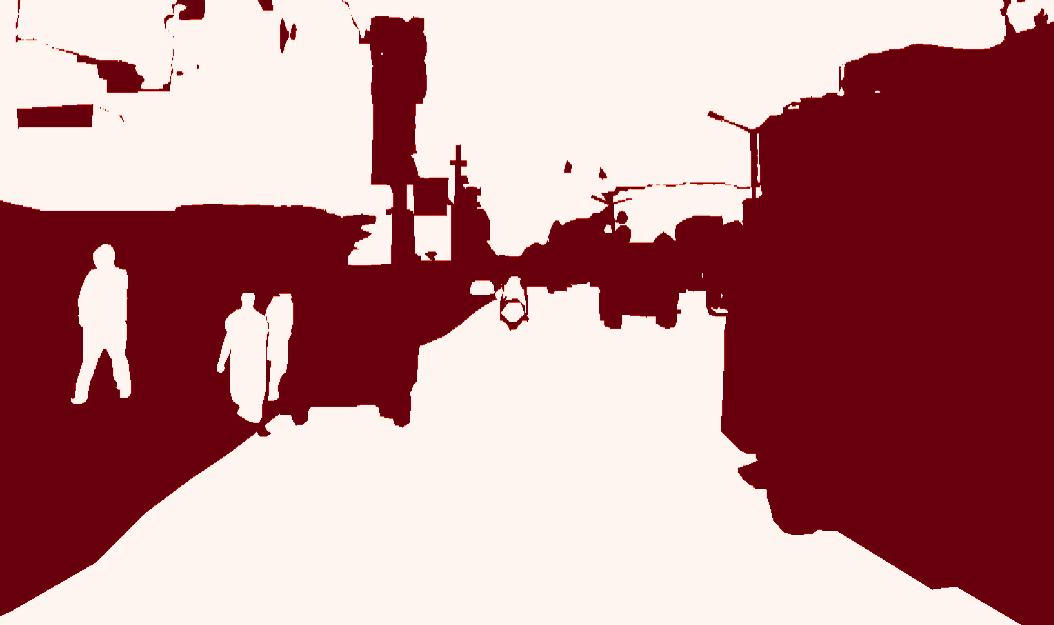}} \\
\end{tabular}
}
\caption{Qualitative results of uncertainty-aware panoptic segmentation by EvPSNet on the Indian Driving data. The visualizations showcase the panoptic segmentation results, corresponding uncertainty, and error maps. The brighter regions in the uncertainty map depict high uncertainty, and the red regions in the error map depict the misclassified pixels by the network.}
\label{fig:idd_ood}
\vspace{-0.4cm}
\end{figure*}

Finally, we present the exciting results of how the network deals with unknown and out-of-distribution scenarios.
For this experiment, we train the network on the Cityscapes dataset and predict on the different Indian Driving dataset.
From the images in \figref{fig:idd_ood}, it is evident that the environment is different in semantic cues like road structure, road dividers, and billboards and instances like auto-rickshaws and different type of vehicles.
In \figref{fig:idd_ood}(a), the panoptic segmentation showcase that the network fails to detect the drivable area due to a lack of semantic boundaries to differentiate between the non-drivable region and the road.
Interestingly though, the uncertainty map predicts high uncertainty for the non-drivable regions.
Thus, uncertainties can be incorporated for robust path planning in the unknown regions.
Similarly, \figref{fig:idd_ood}(b) shows the assignment of high uncertainty to regions like power lines and road dividers.

Moreover, a perception system can encounter novel traffic participants that a network has not encountered during training.
For example, in \figref{fig:idd_ood}(c) and (d), there are auto-rickshaws and a school bus that are missing in the Cityscapes dataset.
Our network predicts high uncertainty for both objects, showcasing the capability to detect out-of-the-distribution objects.
Please note that in all images, the uncertainty still largely correlates with the error map, even though the scenarios are novel for the network.

\end{document}